\documentclass[10pt,twocolumn,letterpaper]{article}

\usepackage[pagenumbers]{cvpr}

\definecolor{cvprblue}{rgb}{0.21,0.49,0.74}
\usepackage[pagebackref,breaklinks,colorlinks,allcolors=cvprblue]{hyperref}
\usepackage{booktabs}
\usepackage{tabularx}
\usepackage{multirow}
\usepackage{graphicx}
\usepackage{cuted}
\usepackage{caption}
\usepackage{stfloats}

\title{D\'ej\`a Cue: Localizing States in Object Histories \\ via Vocabulary-Relative Coordinates}

\author{Haofan Cao\textsuperscript{\rm 1,\rm 2}~~
    Zhichao You\textsuperscript{\rm 1}\thanks{Corresponding author.}~~
    Yunkai Yang\textsuperscript{\rm 1,\rm 2}~~
    Liang Guo\textsuperscript{\rm 1}~~
    Jie Wang\textsuperscript{\rm 1}~~
    Chongshou Li\textsuperscript{\rm 1}
\\
    \textsuperscript{\rm 1}Southwest Jiaotong University, Chengdu, China \quad
    \textsuperscript{\rm 2}University of Leeds, Leeds, UK
}

\begin{document}
\maketitle
\begin{abstract}
Object tracking preserves identity through visual changes, but cannot tell when an object is whole, cut, or peeled. We study state-moment retrieval: locating when a tracked object exhibits each state in a supplied set of descriptions. However, absolute image--text similarity can favor shared object semantics, obscuring the distinctions needed to localize individual states. Our insight is that the alternatives themselves provide a reference for measuring state evidence. We introduce D\'ej\`a Cue, a training-free framework that centers description embeddings on a state-balanced vocabulary mean, producing query directions that compare each state against its alternatives. Using frozen vision--language encoders, D\'ej\`a Cue calibrates frame scores and retrieves intervals at multiple durations within contiguous visible runs. We construct MS-VOST by annotating 620 object histories with 2,465 states. With the encoder and temporal search fixed, D\'ej\`a Cue improves all three localization metrics over absolute scoring across eight encoder configurations, raising Top-1 temporal IoU by 0.42--1.14 percentage points. These gains persist across four text constructions, and candidate-ranking analysis shows that the vocabulary reference brings relevant intervals forward within the same search space. These results establish the supplied state vocabulary as an object-specific reference for localizing semantic states from frozen visual representations.
\end{abstract}    
\section{Introduction}
\label{sec:intro}

\begin{figure}[t]
\centering
\includegraphics[
  width=0.94\columnwidth,
  page=1,
  pagebox=cropbox
]{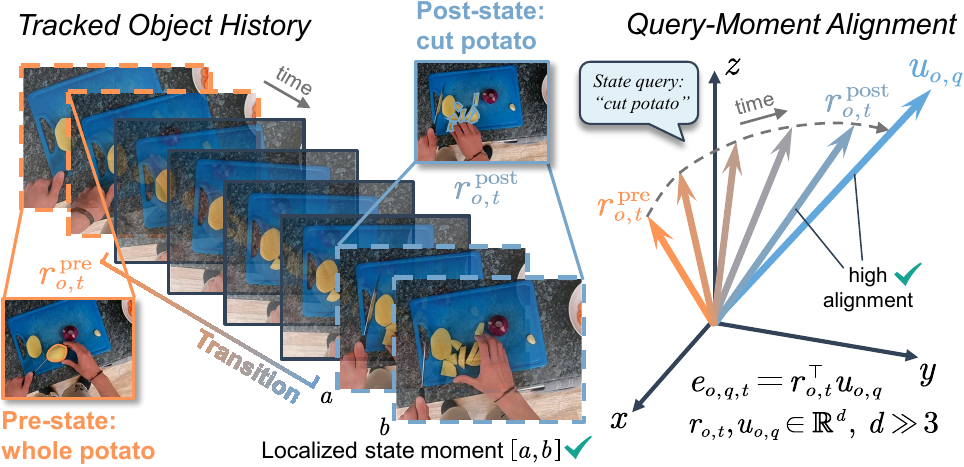}
\vspace{-0.2em}
\caption{Identity-conditioned state-moment retrieval localizes $[a,b]$, where the tracked target exhibits query $q$ (``\emph{cut potato}''); the projection depicts framewise query--moment alignment.}
\vspace{-1.1em}
\label{fig:teaser}
\end{figure}

Persistent visual representations organize video into object histories, yet identity continuity alone does not reveal when each semantic state holds. We study \emph{identity-conditioned state-moment retrieval}: given a tracked-object history and a closed, state-grouped vocabulary of sibling descriptions, localize one temporal interval for each description (Figure~\ref{fig:teaser}). Object association fixes what persists, the supplied sibling set specifies the state distinctions that retrieval must resolve.

Absolute image--text similarity conflates state distinctions with compatibility shared by all siblings. A conventional readout scores each description independently even though siblings repeat category, material, and context terms and every frame depicts the same tracked identity. Let $\bar z_{o,q}$ be the normalized embedding of query $q$ and $c_o$ the state-balanced vocabulary centroid. For a normalized visual feature $y_t$,
\begin{equation}
y_t^\top\bar z_{o,q}=y_t^\top(\bar z_{o,q}-c_o)+y_t^\top c_o.
\label{eq:intro-decomposition}
\end{equation}
The first term measures support for $q$ above the sibling average; the second is shared by every sibling at frame $t$. Temporal variation in the shared term can elevate a broadly object-compatible frame regardless of the state it depicts. The vocabulary therefore supplies an object-specific contrast origin, converting a list of queries into the reference needed for state discrimination.

Because the origin depends on the sibling set, one description can induce different directions under different alternatives. Raw-versus-cut and peeled-versus-unpeeled vocabularies request distinct visual contrasts for the same potato. The coordinate represents this object-specific distinction at inference time; a corpus-level origin cannot.

\textbf{D\'ej\`a Cue} realizes this contrast by averaging descriptions within each state, weighting states equally in $c_o$, and normalizing each residual $\bar z_{o,q}-c_o$ into a vocabulary-relative query direction (Figure~\ref{fig:method}). This coordinate is invariant to a common embedding translation and exact within-state duplication. Median--MAD calibration then rescales frame responses, and a run-constrained multiscale scan retrieves unknown-duration intervals while respecting visibility gaps. Matched coordinate variants keep other component fixed, identifying the origin responsible for the retrieval gain.

We construct MS-VOST dataset by manually annotating videos from VOST~\citep{tokmakov2023vost} with object states, descriptions, and temporal intervals, yielding 620 object histories and 2,465 states. Under the same temporal search, Déjà Cue improves localization across four encoder families and four SigLIP2 model sizes, with Top-1 temporal IoU gains of 0.42–1.14 percentage points. Candidate-ranking analyses show that the coordinate change promotes useful intervals within the same search space. Vocabulary-relative inputs also improve learning-based temporal localization models.

Our contributions are:
\begin{itemize}
\item We formulate state-moment retrieval over persistent object histories, where a complete sibling vocabulary defines the distinctions to localize.
\item We introduce D\'ej\`a Cue, which centers each query on a state-balanced sibling vocabulary and localizes its evidence inside observed runs, using frozen features and no task-specific training.
\item We establish the query-origin effect on MS-VOST, our manually annotated dataset of 620 object histories, with consistent localization gains across four encoder families and four SigLIP2 model sizes.
\end{itemize}
\section{Related Work}

\paragraph{Object states and persistent histories.}
Persistent object histories require both semantic change and identity continuity. Object-state methods discover state-changing actions \citep{soucek2022changeit}, preserve identity through transformations \citep{yu2023vscos}, recognize open-world changes \citep{xue2024vidosc}, predict interaction-induced changes \citep{zameni2025moscato}, or segment spatially progressing transformations \citep{mandikal2026spoc}. Dynamic scene models maintain content through deformation and topology change \citep{park2021hypernerf,luiten2024dynamic3dgaussians}, while language fields attach open-vocabulary semantics to static and dynamic scenes \citep{kerr2023lerf,qin2024langsplat,li20254dlangsplat,xu2026langfield4d}. We take a tracked history as input and retrieve when each member of its jointly supplied state vocabulary holds.

\paragraph{Temporal grounding.}
Temporal language grounding provides the closest localization machinery. Supervised methods learn proposal interactions and boundaries \citep{lei2021momentdetr,jang2023eatr,lin2023univtg,yan2023unloc,xiao2024uvcom,sun2024trdetr,yang2024taskweave}, with query-dependent and object-centric variants conditioning on a sentence or relevant tracklets \citep{moon2023qddetr,moon2026cgdetr,moon2026cva,li2026objectcentricvmr}. Scalable and multimodal systems refine the search space or enrich query semantics \citep{mu2024snag,sun2025momentquant,tang2025simdetr,an2026hieramamba,pramanick2025enrich,zhang2026timelens,halbe2026verve}, while training-free approaches combine frozen priors, captions, or multimodal language models with proposal scoring and temporal postprocessing \citep{luo2024zeroshotvmr,zheng2024tfvtg,xu2025momentgpt,jeon2026granalign,diwan2023zeroshot,jeon2025pointtospan}. These methods generally encode each sentence independently; identity-conditioned retrieval instead uses the complete sibling set to define every query coordinate.

\paragraph{Relative embedding geometry.}
Embedding corrections derive reference origins from corpus means or dominant directions \citep{mu2018allbuttop}, negative-set statistics, broad concept vocabularies, or modality-level geometry \citep{zhou2023distribution,bhalla2024splice,levi2025doubleellipsoid}. Such references are global and independent of the state alternatives supplied for a particular object. D\'ej\`a Cue uses those alternatives as conditioning context, forming a state-balanced origin for each history so that every score measures support for $q$ relative to its siblings. It requires no cross-history statistics and is invariant to exact within-state duplication.
\section{Problem Formulation}

Identity-conditioned state-moment retrieval maps a tracked-object history and a closed, state-grouped sibling vocabulary to one interval per description. For object history $o$, let $x_{o,t}\in\mathbb{R}^{d}$ be the tracked-object embedding at frame $t$, $v_{o,t}\in\{0,1\}$ indicate an observed target, and $z_{o,q}\in\mathbb{R}^{d}$ encode description $q\in\{1,\ldots,Q_o\}$. The supplied partition $\Pi_o=\{\mathcal Q_{o,s}\}_{s=1}^{S_o}$ groups descriptions into $S_o\geq2$ sibling states. Visual and text embeddings share a frozen encoder space. The input is
\begin{equation}
\begin{aligned}
X_o&=\{x_{o,t}\}_{t=1}^{T},\quad
V_o=\{v_{o,t}\}_{t=1}^{T},\\
Z_o&=\{z_{o,q}\}_{q=1}^{Q_o},\quad
\Pi_o=\{\mathcal Q_{o,s}\}_{s=1}^{S_o},
\end{aligned}
\end{equation}
Here $\Pi_o$ records the grouping of descriptions into sibling states. Frame labels, reference intervals, occurrence order, and event anchors remain unknown. The output is an inclusive window $\hat w_{o,q}=[\hat a_{o,q},\hat b_{o,q}]$ for every $q$.

Each output represents one occurrence of the queried state. All queries share the object association and complete vocabulary $Z_o$. Frames with $v_{o,t}=0$ retain their temporal indices but contribute no evidence. An \emph{observed run} is a maximal contiguous interval with $v_{o,t}=1$, and every prediction lies within one run.
\section{Method}

\begin{figure*}[t]
\centering
\includegraphics[width=\textwidth,pagebox=cropbox]{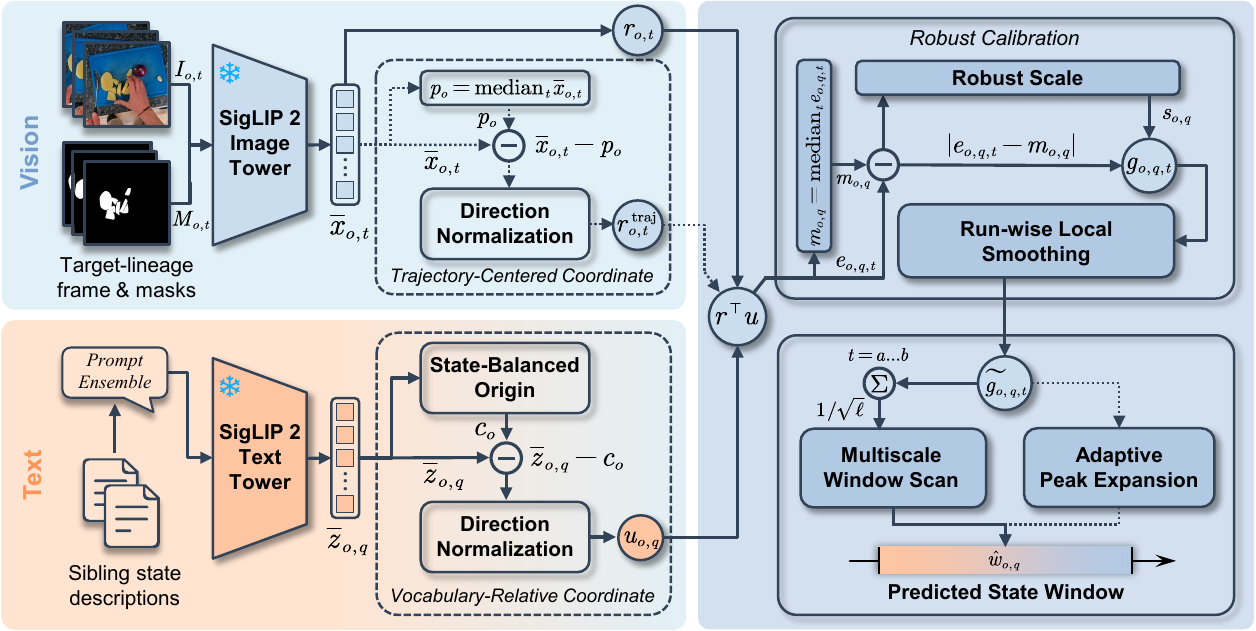}
\caption{\textbf{Overview.} Each description is expressed relative to the state-balanced sibling centroid and scored against target-lineage frames. Robust scaling and run-wise smoothing calibrate the responses before a multiscale search selects a window within one observed run.}
\vspace{-1.5em}
\label{fig:method}
\end{figure*}

D\'ej\`a Cue first expresses each description against its state-balanced siblings, then calibrates the resulting frame evidence and searches multiple durations within observed runs (Figure~\ref{fig:method}). The first stage changes the semantic reference; the second resolves response scale, visibility gaps, and unknown duration without changing that reference. 
Supplementary \Cref{sec:supp-method-derivations} derives the coordinate and scan properties.

\subsection{Vocabulary-Relative State Coordinates}

Unequal numbers of paraphrases should not give one sibling state greater influence over the query origin. We first $\ell_2$-normalize every observed frame and description:
\begin{equation}
\bar x_{o,t}=\frac{x_{o,t}}{\|x_{o,t}\|_2},\qquad
\bar z_{o,q}=\frac{z_{o,q}}{\|z_{o,q}\|_2}.
\end{equation}
We then average within each state before averaging the state prototypes:
\begin{equation}
\mu^z_{o,s}=\frac{1}{|\mathcal Q_{o,s}|}
\sum_{q\in\mathcal Q_{o,s}}\bar z_{o,q},\qquad
c_o=\frac{1}{S_o}\sum_{s=1}^{S_o}\mu^z_{o,s}.
\end{equation}
This two-level average assigns every state weight $1/S_o$, independent of its number of paraphrases. 
Adding a state changes the requested distinction, whereas uniformly repeating all descriptions within one state leaves its prototype and the reference unchanged.
Each description is expressed relative to this state-balanced centroid,
\begin{equation}
\widetilde u_{o,q}=\bar z_{o,q}-c_o,
\end{equation}
and normalized into a query direction:
\begin{equation}
u_{o,q}=\begin{cases}
\widetilde u_{o,q}/\|\widetilde u_{o,q}\|_2,
& \|\widetilde u_{o,q}\|_2>\tau,\\
\bar z_{o,q}, & \text{otherwise}.
\end{cases}
\end{equation}
We set $\tau=10^{-8}$; the fallback covers a query that is indistinguishable from its sibling centroid.

State-balanced projection removes a shared translation exactly. Let $\bar M_o=[\mu^z_{o,1},\ldots,\mu^z_{o,S_o}]$ and $P_o=I-S_o^{-1}\mathbf 1\mathbf 1^\top$. The centered prototypes $G_o=\bar M_oP_o$ are invariant to any common offset $\delta\in\mathbb R^d$ because
\begin{equation}
(\bar M_o+\delta\mathbf 1^\top)P_o=\bar M_oP_o.
\end{equation}
The projection is zero-sum and preserves pairwise description differences before normalization. For any visual vector $y$, define $a_{o,q}(y)=y^\top\bar z_{o,q}$. Then
\begin{equation}
y^\top\widetilde u_{o,q}=a_{o,q}(y)
-\frac{1}{S_o}\sum_{s=1}^{S_o}\frac{1}{|\mathcal Q_{o,s}|}
\sum_{j\in\mathcal Q_{o,s}}a_{o,j}(y).
\label{eq:8}
\end{equation}
The score is therefore similarity to $q$ relative to the state-balanced sibling average. A frame ranks highly when it favors $q$ over that average; compatibility shared by every sibling cancels. Direction normalization preserves this contrast orientation while equalizing residual magnitudes.

\begin{figure}[htbp]
    \centering
    \includegraphics[width=1.0\linewidth]{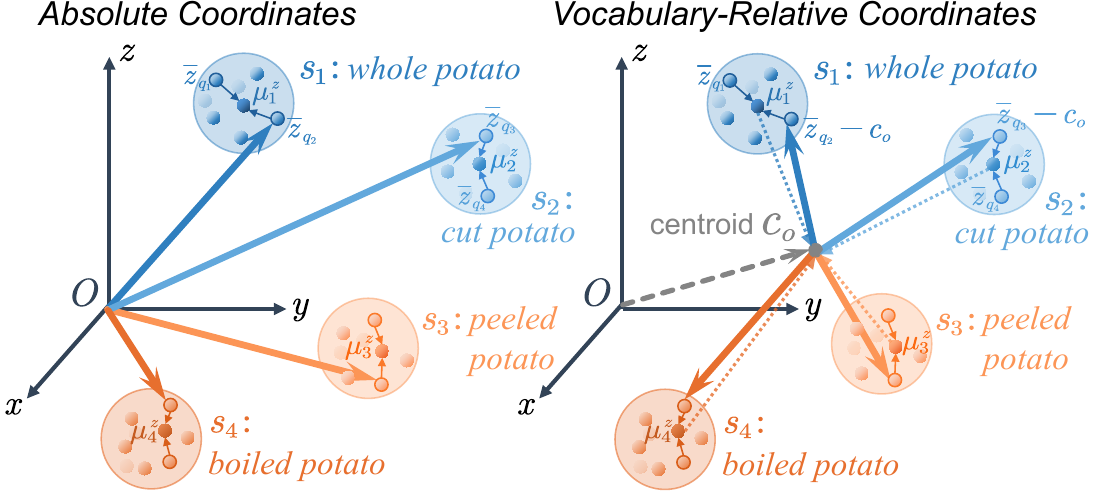}
    \caption{\textbf{Geometry intuition.} Absolute queries retain object and context compatibility shared across sibling states, whereas Déjà Cue subtracts the state-balanced vocabulary centroid \(c_o\), exposing state-specific contrast directions.}
    \vspace{-1em}
    \label{fig:intuition}
\end{figure}

The geometry is especially transparent for two states. Their centered prototypes are opposite directions, $\mu^z_{o,1}-c_o=(\mu^z_{o,1}-\mu^z_{o,2})/2$ and $\mu^z_{o,2}-c_o=-(\mu^z_{o,1}-\mu^z_{o,2})/2$, so evidence is read along the semantic axis joining the alternatives. With more states, each prototype is contrasted against their barycenter (\Cref{fig:intuition}). Description-level residuals preserve paraphrase-specific variation around these directions.

Frame evidence uses normalized visual features directly:
\begin{equation}
r_{o,t}=\bar x_{o,t},\qquad
e_{o,q,t}=r_{o,t}^{\top}u_{o,q}.
\end{equation}
For a complementary history-relative visual coordinate, we subtract the coordinate-wise trajectory median
\begin{align}
p_o&=\operatorname*{median}_{t:v_{o,t}=1}\bar x_{o,t},\nonumber\\
r^{\mathrm{traj}}_{o,t}&=\frac{\bar x_{o,t}-p_o}
 {\max(\|\bar x_{o,t}-p_o\|_2,10^{-12})}.
\end{align}
The dual variant retains $u_{o,q}$ and substitutes
$r^{\mathrm{traj}}_{o,t}$, giving
\begin{equation}
e^{\mathrm{dual}}_{o,q,t}=(r^{\mathrm{traj}}_{o,t})^{\top}u_{o,q}.
\end{equation}
The trajectory-only ablation instead pairs $r^{\mathrm{traj}}_{o,t}$ with
$\bar z_{o,q}$, while the matched absolute baseline uses
$(\bar x_{o,t},\bar z_{o,q})$. These four coordinate systems differ only in their
visual or query origin.

\subsection{Run-Constrained Temporal Localization}
\label{sec:4.2}
Temporal localization must accommodate query-specific score scales, visibility gaps, and unknown state duration. Median--MAD standardization first calibrates each response over observed frames:
\begin{align}
m_{o,q}&=\operatorname*{median}_{t:v_{o,t}=1}e_{o,q,t},\\
s_{o,q}&=\max\!\left(1.4826\operatorname*{median}_{t:v_{o,t}=1}
|e_{o,q,t}-m_{o,q}|,10^{-3}\right),\\
g_{o,q,t}&=(e_{o,q,t}-m_{o,q})/s_{o,q}.\nonumber
\end{align}
The median defines the response of a typical observed frame, while the MAD measures its robust temporal spread. Positive standardized values therefore indicate evidence above the history-specific background without allowing a few extreme frames to determine the scale.
When the MAD floor is inactive, this calibration is invariant to every positive affine transformation $e'_{o,q,t}=\alpha e_{o,q,t}+\beta$, $\alpha>0$: the median becomes $\alpha m_{o,q}+\beta$, while the MAD and centered residual both scale by $\alpha$, giving $g'_{o,q,t}=g_{o,q,t}$. Window ranking therefore depends on the temporal response pattern, not its offset or positive scale.
The $10^{-3}$ floor prevents unstable division for a nearly constant response; the $\tau$ branch analogously retains the absolute query when sibling contrast vanishes.
Local smoothing then aggregates support independently inside each observed run. For an odd support width $k$, let $r_k=(k-1)/2$ and $J_R^{(k)}(t)=\{j\in\mathbb Z:|j|\le r_k,\ t+j\in R\}$ for frame $t$ in run $R$, with $h_j=1/k$:
\begin{equation}
\widetilde g_{o,q,t}=
\frac{\sum_{j\in J_R^{(k)}(t)}h_j g_{o,q,t+j}}
{\sqrt{\sum_{j\in J_R^{(k)}(t)}h_j^2}}.
\end{equation}
Support contracts at run boundaries, and active-tap normalization keeps its noise scale comparable. Under independent equal-variance frame noise, the denominator is the active filter's standard deviation; shortened boundary kernels therefore remain comparable with interior kernels. Evidence next to a visibility gap can compete without mixing observations across that gap.

\vspace{-1em}

\paragraph{Multiscale search.}
To handle unknown state duration, a geometric $1.5\times$ schedule sets candidate lengths; every fitting length is enumerated at every start within each observed run. A window $w=[a,b]$ of length $\ell=b-a+1$ receives
\begin{equation}
S(o,q,w)=\frac{1}{\sqrt{\ell}}\sum_{t=a}^{b}\widetilde g_{o,q,t}.
\end{equation}
The prediction is $\hat w_{o,q}=\arg\max_w S(o,q,w)$, with ties resolved by earliest start and then earliest end. If positive evidence of mean $\mu$ persists for $L$ frames, the expected score grows as $\mu\sqrt{\ell}$ up to $L$ and decays as $\mu L/\sqrt{\ell}$ beyond it. Prefix sums evaluate schedule in $O(T\log T)$ time per query.

Calibration removes query-specific offset and scale, observed runs enforce visibility, and the geometric schedule accommodates unknown duration. None introduces a learned boundary prior or temporal supervision.
\section{Experiments}
\label{sec:experiments}

Our experiments ask which reference helps localize states, how that reference changes temporal interval selection, and whether learned models can use it. We address these questions on MS-VOST through coordinate comparisons, candidate-ranking analysis, and vocabulary-relative inputs.

\begin{table*}[!t]
\centering
\caption{\textbf{Training-free temporal localization on MS-VOST
across encoder families and SigLIP2 scales.}
Each row compares Absolute and D\'ej\`a Cue using the same encoder
and temporal search.}
\label{tab:encoder-comparison}
\small
\setlength{\tabcolsep}{3pt}
\resizebox{\linewidth}{!}{%
\begin{tabular}{@{}cllrr*{3}{ccr}@{}}
\toprule
& & & & &
\multicolumn{3}{c}{R@1@0.3 $\uparrow$} &
\multicolumn{3}{c}{R@1@0.5 $\uparrow$} &
\multicolumn{3}{c}{Top-1 tIoU $\uparrow$} \\
\cmidrule(lr){6-8}
\cmidrule(lr){9-11}
\cmidrule(lr){12-14}
& Encoder & Scale & Pixels & Dim
& Absolute & D\'ej\`a Cue & $\Delta$
& Absolute & D\'ej\`a Cue & $\Delta$
& Absolute & D\'ej\`a Cue & $\Delta$ \\
\midrule
\multirow{4}{*}{\rotatebox[origin=c]{90}{\shortstack{Encoder\\families}}}
& CLIP \citep{radford2021clip} & Base & 224 & 512
& $13.34$ & $14.40$~ & \textbf{+1.06}
& $6.31$ & $6.94$~ & \textbf{+0.64}
& $9.90$ & $10.59$~ & \textbf{+0.68} \\

& SigLIP1 \citep{zhai2023siglip} & Base & 224 & 768
& $13.76$ & $14.44$~ & \textbf{+0.68}
& $6.42$ & $6.82$~ & \textbf{+0.40}
& $10.17$ & $10.59$~ & \textbf{+0.42} \\

& MetaCLIP \citep{xu2024metaclip} & Base & 224 & 512
& $13.47$ & $14.73$~ & \textbf{+1.26}
& $6.26$ & $6.89$ &~ \textbf{+0.63}
& $10.12$ & $10.95$~ & \textbf{+0.82} \\

& SigLIP2 \citep{tschannen2025siglip2} & Base & 224 & 768
& $13.27$ & $14.54$~ & \textbf{+1.27}
& $6.44$ & $7.14$~ & \textbf{+0.69}
& $10.01$ & $10.78$~ & \textbf{+0.78} \\

\midrule
\multirow{4}{*}{\rotatebox[origin=c]{90}{\shortstack{SigLIP2\\scales}}}
& SigLIP2 \citep{tschannen2025siglip2} & Base & 256 & 768
& $13.28$ & $14.38$~ & \textbf{+1.10}
& $6.58$ & $7.06$~ & \textbf{+0.49}
& $10.09$ & $10.82$~ & \textbf{+0.73} \\

& SigLIP2 \citep{tschannen2025siglip2} & Large & 256 & 1024
& $13.08$ & $14.82$~ & \textbf{+1.74}
& $6.17$ & $7.00$~ & \textbf{+0.82}
& $9.98$ & $10.82$~ & \textbf{+0.84} \\

& SigLIP2 \citep{tschannen2025siglip2} & So400m & 256 & 1152
& $13.88$ & $14.88$~ & \textbf{+1.00}
& $6.30$ & $7.27$~ & \textbf{+0.97}
& $10.14$ & $11.27$~ & \textbf{+1.14} \\

& SigLIP2 \citep{tschannen2025siglip2} & Giant-opt & 256 & 1536
& $13.24$ & $15.26$~ & \textbf{+2.02}
& $6.42$ & $7.44$~ & \textbf{+1.01}
& $9.93$ & $11.04$~ & \textbf{+1.12} \\

\bottomrule
\end{tabular}%
}
\par\vspace{3pt}
\begin{minipage}{\linewidth}
\footnotesize
\textit{Note.} Absolute uses uncentered embeddings; D\'ej\`a Cue (ours) centers only text
embeddings on the state-balanced sibling-vocabulary mean.
Scores are percentages; $\Delta$ is D\'ej\`a Cue minus Absolute (pp).
Pixels denotes input resolution; Dim denotes embedding dimension. All encoders use $16 \times 16$ patches.
\end{minipage}
\vspace{-1em}
\end{table*}

\subsection{Experimental Setup}

\paragraph{Dataset.} We construct MS-VOST (Multi-State VOST) dataset by manually annotating VOST~\citep{tokmakov2023vost} videos with object states, state descriptions, and temporal intervals (two annotators independently label every frame), yielding 620 object histories. These histories contain 2,465 states, 4,930 descriptions, 2,473 annotated intervals, and 582,705 video frames. Each history contains three to five states, and each state's label and visual definition serve as separate queries. 

\vspace{-1em}

\paragraph{Features and queries.} All methods share target masks, object crops, state descriptions, and visibility information. Each visible native frame is cropped around the target-mask union with 20\% padding, and pixels outside the mask are set to RGB 127. Image and text encoders remain frozen and produce unit-normalized embeddings. For training-free retrieval, the default text feature is obtained by averaging and renormalizing the embeddings of three prompt forms: the raw description, ``\textit{a photo of} [description]," and ``\textit{the} [description]." SigLIP2 Base/16 at 224 pixels is the reference encoder for the component and learned-model studies.

\paragraph{Metrics.} Temporal IoU (tIoU) measures the overlap between predicted and reference intervals with inclusive endpoints. For each prediction, tIoU is the maximum IoU with the intervals of the queried state. R@1 at thresholds 0.3 and 0.5 averages the indicator that the top-ranked prediction reaches the threshold; Top-1 tIoU averages the unthresholded overlap. For each metric, we average over descriptions within states, states within histories, and histories within source videos, then average uniformly over videos.

\vspace{-1em}

\paragraph{Coordinates comparisons.} Coordinate comparisons share median-MAD calibration, three-frame smoothing, candidate windows, square-root-normalized summed window scores, and tie breaking; then only the coordinate origin varies. Window lengths follow a $1.5\times$ geometric schedule up to the video's native length, with every valid start position evaluated within each contiguous visible run. 
Dataset construction, implementation, and learned-input details appear in Supplementary Sections~\ref{supp:data}, \ref{supp:implementation}, and \ref{supp:learned}.

\subsection{Which reference helps localization?}
\label{sec:which}

\begin{figure}[!h]
    \centering
    \includegraphics[width=0.88\linewidth]{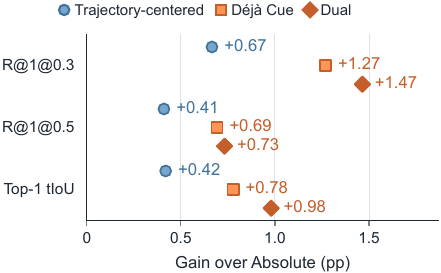}
    \caption{\textbf{Localization gains from visual and vocabulary references.} Dual combines trajectory and vocabulary centering.}
    \vspace{-0.8em}
    \label{fig:reference}
\end{figure}

The sibling vocabulary provides a useful semantic reference for distinguishing states within an object's history. We compare absolute coordinates, a visual reference derived by centering the object's trajectory, and the vocabulary reference used by D\'ej\`a Cue. Since the encoder features, calibration, candidate intervals, and temporal scoring remain constant, this comparison isolates the effect of the reference definition. \Cref{fig:reference} shows that, using the SigLIP2 Base/16 encoder, both references improve localization over absolute coordinates. Vocabulary centering yields the highest performance among these three choices, with R@1 at tIoU 0.5 increasing from 6.44\% to 7.14\%, and Top-1 tIoU rising from 10.01\% to 10.78\%. The visual reference achieves 10.43\% Top-1 tIoU. Thus, the provided descriptions offer an effective way to define the distinctions used to read the fixed visual representation.

\paragraph{State balance.}
Equal state weighting makes the reference reflect the represented states. We compute this by averaging descriptions within each state and then averaging the resulting state prototypes. Consequently, uniformly repeating every description of a single state leaves both its prototype and the reference unchanged. In the dual extension, this construction preserves all 4,930 original-query windows across eight encoders when the two descriptions of the first state are repeated two, four, or eight times. Assigning equal weights to descriptions alters 1,965, 3,064, and 3,640 windows for reference encoder, as detailed in Supplementary \Cref{sec:11.1}. The comparison connects the semantic unit used in the centroid to the stability of interval selection.

\vspace{-1em}

\paragraph{Across encoders.}
\Cref{tab:encoder-comparison} compares four Base/16 encoder families at 224 pixels and four
SigLIP2 model sizes at 256 pixels, all with $16 \times 16$ patches.
With the encoder and temporal search fixed within each comparison,
D\'ej\`a Cue improves all three localization metrics across all eight
configurations, with Top-1 tIoU gains of 0.42--1.14 percentage points.
The vocabulary reference therefore remains useful across the tested
families and scales, complementing the choice of representation.

\begin{figure}[t]
\centering
\includegraphics[width=0.84\columnwidth]{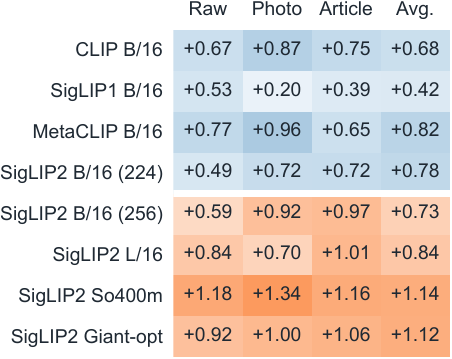}
\vspace{-0.2em}
\caption{\textbf{Vocabulary centering across text inputs on MS-VOST.} Cells report Top-1 tIoU gains (D\'ej\`a Cue minus Absolute, percentage points). \textit{Raw}, \textit{Photo}, and \textit{Article} encode $d$, ``\textit{a photo of} $d$,'' and ``\textit{the} $d$''; \textit{Avg.} averages and renormalizes their embeddings.}
\vspace{-1em}
\label{fig:text-inputs}
\end{figure}

\paragraph{Does the gain depend on text templates?} The cross-encoder comparison averages 3 text forms for each description. To test whether the gain depends on this ensemble, we evaluate the same coordinate change under each form individually---the raw description $d$, ``\textit{a photo of} $d$'', and ``\textit{the} $d$''---alongside their normalized average. Within each encoder and text-form pair, Absolute and D\'ej\`a Cue share identical visual features and temporal search. D\'ej\`a Cue improves all localization metrics across all 32 encoder--input configurations, raising Top-1 tIoU by 0.20--1.34 percentage points (Figure~\ref{fig:text-inputs}). Because these gains hold for single text forms as well as the default average, the benefit of centering on sibling descriptions extends across the tested encoder families, model scales, and query-feature constructions.

Together, reference comparison identifies a useful source of context, while the encoder and text-input analyses show that its benefit carries across the tested representations and query constructions. We next examine how changing coordinates affects ranking and selection of temporal intervals.

\subsection{How does reference change interval selection?}

The shared candidate set allows us to distinguish between interval selection and the intervals available to the search. For the reference encoder, the mean best-candidate tIoU remains 84.60\% for both absolute and vocabulary-relative coordinates. We assess the placement of relevant intervals, defined as having a tIoU of at least 0.5, within each ranking. Mean reciprocal rank (MRR) rewards an early first relevant interval, while R@K determines if any relevant interval appears within the top K candidates. These complementary measures relate the top prediction to the ordering of the candidate set.

\begin{figure}[htbp]
    \centering
    \includegraphics[width=0.85\linewidth]{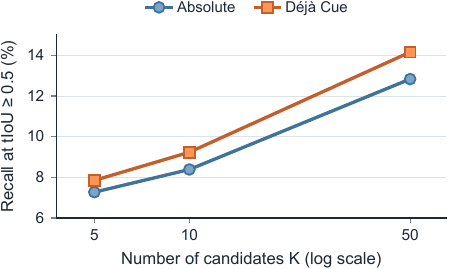}
    \vspace{-0.3em}
    \caption{\textbf{Vocabulary centering improves candidate ranking.}
Top-$K$ recall at tIoU $\geq 0.5$ on MS-VOST with SigLIP2. Absolute and D\'ej\`a Cue rank the same candidate
intervals using uncentered and vocabulary-centered text embeddings,
respectively.}
\vspace{-0.7em}
    \label{fig:ranking}
\end{figure}

The vocabulary reference brings relevant intervals forward in the ranking. MRR increases from 7.15\% to 7.91\%, while R@5, R@10, and R@50 increase from 7.27\%, 8.39\%, and 12.83\% to 7.85\%, 9.24\%, and 14.16\%, respectively (\Cref{fig:ranking}). The gain at each evaluated cutoff shows that the effect appears across several portions of the ranked list. Together with the MRR $\uparrow$, this supports better access to relevant intervals near the front of the same candidate set. The localization improvement in \Cref{sec:which} is thus accompanied by a broader change in candidate ordering.

\begin{figure}[!t]
\centering
\includegraphics[width=\columnwidth,pagebox=cropbox]{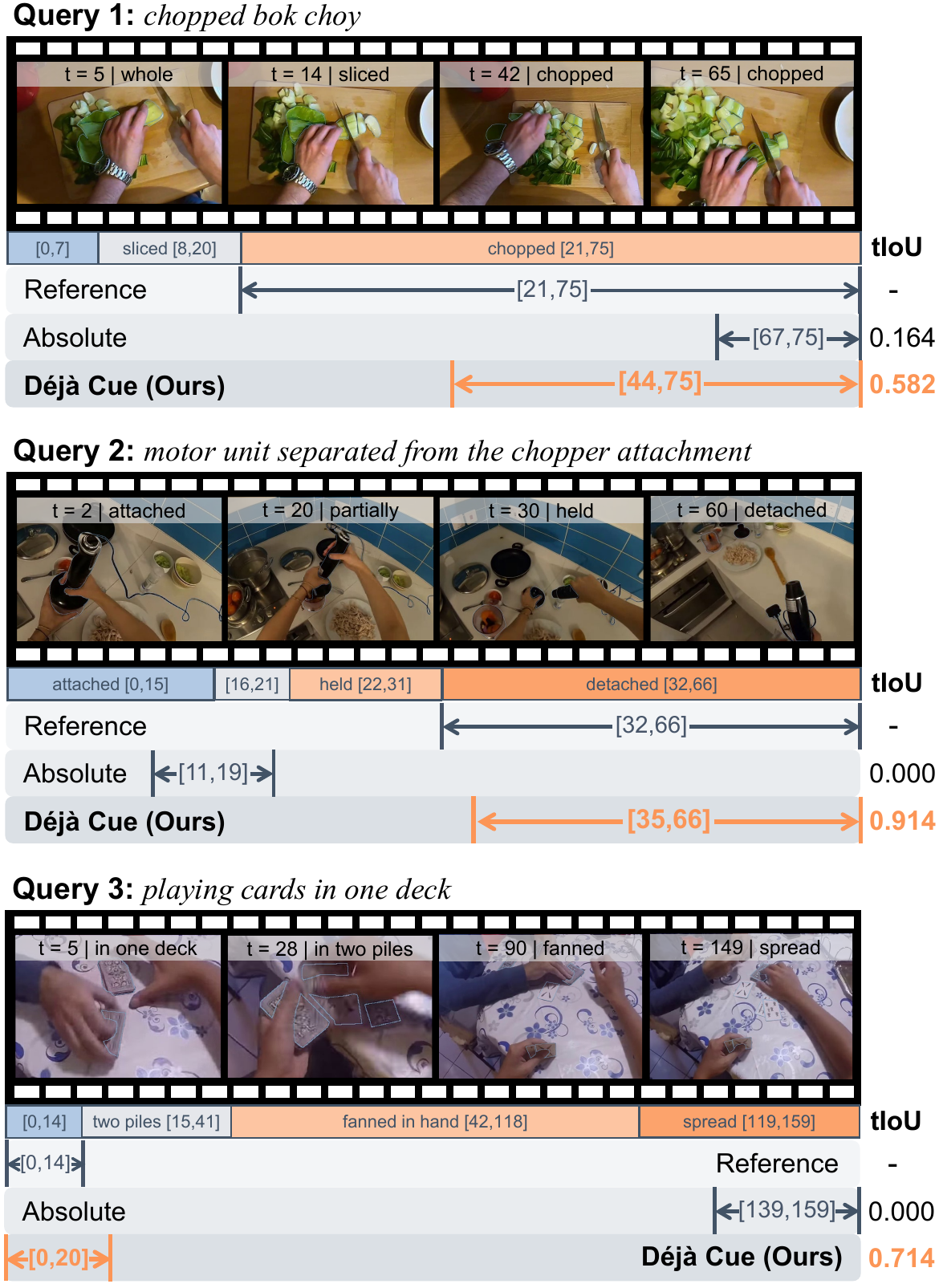}
\caption{\textbf{Qualitative interval selection on MS-VOST.}
Each example shows sampled frames and aligned intervals for
Reference (i.e. ground truth), Absolute, and D\'ej\`a Cue.
Interval endpoints are frame indices; tIoU is shown at right.}
\vspace{-1.5em}
\label{fig:formal-cases}
\end{figure}

\Cref{fig:formal-cases} illustrates how vocabulary-relative queries change interval selection across three multi-state object histories. For ``\textit{chopped bok choy}", Absolute retrieves only the final portion of the annotated state, selecting frames 67--75 against the reference interval of 21--75. D\'ej\`a Cue extends the prediction to frames 44--75, recovering more of the target interval and increasing tIoU from 0.164 to 0.582. For ``\textit{motor unit separated from the chopper attachment}", Absolute selects frames 11--19, before the reference interval of 32--66. D\'ej\`a Cue instead selects frames 35--66, covering 32 of the 35 annotated frames and yielding a tIoU of 0.914, compared with 0 for Absolute method. 

These cases illustrate improved temporal coverage and corrections in both temporal directions. The sampled frames provide the surrounding visual context; the aligned intervals show exactly which part of the history each reading retrieves. These examples connect the aggregate ordering results to the task of identifying when the queried state occurs within an object's observation sequence.

The ordering analysis also clarifies what the vocabulary changes in the score. \Cref{eq:8} expresses each unnormalized response as similarity to the queried description relative to the state-balanced sibling average. The average similarity can vary with the observed frame, so the reference can change the temporal response pattern used to rank intervals. Median--MAD calibration already removes constant score offsets and positive rescaling when its scale floor is inactive (\Cref{sec:4.2}). The query reference thus acts through how individual observations compare with the supplied states. The ranking results and aligned examples show the retrieval consequence of this construction.

\begin{figure}[htbp]
\centering
\includegraphics[width=0.9\columnwidth]{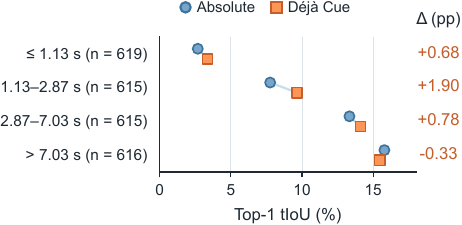}
\vspace{-0.3em}
\caption{\textbf{Localization across state durations on MS-VOST.}
States are grouped into quartiles by mean annotated occurrence duration;
$n$ denotes states per group and $\Delta$ is D\'ej\`a Cue minus Absolute
in percentage points.}
\vspace{-1em}
\label{fig:state-duration}
\end{figure}

\paragraph{How does localization vary with state duration?} Grouping the 2,465 states into dataset quartiles by mean annotated occurrence duration ($n = 619, 615, 615, 616$), we evaluate both coordinates with the reference encoder under the same temporal search (Figure~\ref{fig:state-duration}). Across the first three quartiles, D\'ej\`a Cue raises Top-1 tIoU from 2.68\% to 3.36\%, from 7.76\% to 9.66\%, and from 13.33\% to 14.11\%, corresponding to gains of 0.68, 1.90, and 0.78 percentage points. The largest improvement appears for intermediate durations of 1.13--2.87 seconds, whereas in the longest quartile (above 7.03 seconds) overlap decreases slightly from 15.79\% to 15.45\% ($-0.33$ percentage points). Duration thus distinguishes absolute localization accuracy from the benefit of the reference: the longest states have the highest overlap under both coordinates, whereas vocabulary centering provides its largest gain at an intermediate duration. The positive gains across the first three quartiles place the aggregate improvement across several temporal scales.

\vspace{-1em}

\paragraph{Does the gain depend on boundary placement?} Because gradual state transitions admit more than one valid onset and offset, we rescore the same fixed predictions against the stable reference interval and three annotated alternatives that expand its start, its end, or both endpoints to plausible boundaries. As in the duration analysis, we use SigLIP2 Base/16 at $224\times224$ resolution. D\'ej\`a Cue improves all three localization metrics under all four boundary definitions, with Top-1 tIoU gains of 0.78--0.85 percentage points (Table~\ref{tab:boundary-choices}). The advantage of vocabulary centering therefore holds across annotated interpretations of where a state begins and ends, complementing the duration analysis with a direct check on reference-interval definitions.

\begin{table}[htbp]
\centering
\caption{\textbf{Localization under annotated boundary choices on MS-VOST.}
Top-1 tIoU (\%) with fixed predictions.
\textit{Earlier start} and \textit{Later end} extend the corresponding
endpoint of the \textit{Stable} interval to its annotated plausible
boundary; \textit{Broader} extends both.}
\label{tab:boundary-choices}
\small
\setlength{\tabcolsep}{5pt}
\begin{tabular}{@{}lccr@{}}
\toprule
Boundary & Absolute & D\'ej\`a Cue & $\Delta$ \\
\midrule
Stable & $10.01$ & $10.78$ & \textbf{+0.78} \\
Earlier start & $10.29$ & $11.11$ & \textbf{+0.82} \\
Later end & $10.04$ & $10.85$ & \textbf{+0.81} \\
Broader & $10.33$ & $11.18$ & \textbf{+0.85} \\
\bottomrule
\end{tabular}
\vspace{-1em}
\end{table}

\begin{table*}[!t]
\centering
\caption{\textbf{Effect of vocabulary-relative coordinates on learning-based temporal localization models on MS-VOST.}}
\vspace{-0.4em}
\label{tab:learned-vocabulary-coordinates}
\small
\setlength{\tabcolsep}{3pt}
\resizebox{\linewidth}{!}{%
\begin{tabular}{@{}ll*{9}{r}@{}}
\toprule
& & \multicolumn{3}{c}{R@1@0.3 $\uparrow$} & \multicolumn{3}{c}{R@1@0.5 $\uparrow$} & \multicolumn{3}{c}{Top-1 tIoU $\uparrow$} \\
\cmidrule(lr){3-5}\cmidrule(lr){6-8}\cmidrule(lr){9-11}
Method & Venue & Original & VRC & $\Delta$ & Original & VRC & $\Delta$ & Original & VRC & $\Delta$ \\
\midrule
Moment-DETR \citep{lei2021momentdetr} & NeurIPS 2021 & $33.44_{\pm 0.38}$ & $35.40_{\pm 0.81}$ & \textbf{+1.96} & $21.36_{\pm 0.60}$ & $23.82_{\pm 0.74}$ & \textbf{+2.46} & $23.42_{\pm 0.28}$ & $25.29_{\pm 0.41}$ & \textbf{+1.87} \\
QD-DETR \citep{moon2023qddetr} & CVPR 2023 & $32.89_{\pm 0.10}$ & $39.12_{\pm 1.30}$ & \textbf{+6.23} & $21.39_{\pm 0.57}$ & $25.85_{\pm 0.88}$ & \textbf{+4.46} & $23.59_{\pm 0.29}$ & $27.86_{\pm 0.84}$ & \textbf{+4.27} \\
UVCOM \citep{xiao2024uvcom} & CVPR 2024 & $34.04_{\pm 0.81}$ & $41.40_{\pm 0.10}$ & \textbf{+7.36} & $22.13_{\pm 0.29}$ & $28.50_{\pm 0.70}$ & \textbf{+6.37} & $24.17_{\pm 0.18}$ & $29.10_{\pm 0.30}$ & \textbf{+4.93} \\
TR-DETR \citep{sun2024trdetr} & AAAI 2024 & $26.88_{\pm 0.67}$ & $25.12_{\pm 0.43}$ & -1.77 & $13.70_{\pm 0.88}$ & $13.38_{\pm 0.81}$ & -0.32 & $19.82_{\pm 0.51}$ & $18.29_{\pm 0.40}$ & -1.54 \\
TaskWeave \citep{yang2024taskweave} & CVPR 2024 & $30.79_{\pm 5.46}$ & $40.31_{\pm 4.98}$ & \textbf{+9.52} & $18.52_{\pm 4.61}$ & $26.89_{\pm 4.73}$ & \textbf{+8.37} & $21.51_{\pm 3.60}$ & $28.64_{\pm 3.33}$ & \textbf{+7.13} \\
Sim-DETR \citep{tang2025simdetr} & ICCV 2025 & $30.31_{\pm 0.60}$ & $27.05_{\pm 1.79}$ & -3.26 & $16.79_{\pm 0.80}$ & $13.55_{\pm 1.15}$ & -3.24 & $22.03_{\pm 0.29}$ & $20.08_{\pm 1.05}$ & -1.95 \\
\bottomrule
\end{tabular}%
}
\par\vspace{3pt}
\begin{minipage}{\linewidth}
\footnotesize
\textit{Note.} Original uses query-token features; VRC uses vocabulary-relative coordinates (input).
All models use frozen SigLIP2 Base/16 visual features.
Scores (\%) are $\mathrm{mean}_{\pm\mathrm{SD}}$ over three seeds, with five-fold
predictions combined within each seed before scoring.
$\Delta$ is VRC minus Original in percentage points.
\end{minipage}
\vspace{-1.2em}
\end{table*}

\subsection{Can learned models use the reference?}
\label{sec:5.4}

The sibling vocabulary also provides context for learned temporal localization. We investigate whether a learned predictor can use a vocabulary-relative language representation alongside the visual sequence. The native input consists of token features from the query sentence. Our vocabulary-relative input includes pooled sibling-description vectors, centered on the state-balanced vocabulary mean and normalized, alongside an indicator for the queried description. Each prediction is thus conditioned on the queried state and its alternatives. We compare these representations within each architecture using fixed SigLIP2 visual features.

We use five folds grouped by source video and 3 random seeds, with 200 training epochs per run (\Cref{supp:learned}). \Cref{tab:learned-vocabulary-coordinates} reports the mean and
standard deviation across seeds after combining the predictions from
all 5 evaluation folds within each seed before scoring. Visual
features and training schedules remain fixed within each comparison.

Vocabulary-relative inputs improve all three metrics for Moment-DETR, QD-DETR, UVCOM and TaskWeave. For these four models, gains in R@1 at tIoU 0.5 range from 2.46 to 8.37 percentage points, and Top-1 tIoU gains range from 1.87 to 7.13 points. The agreement between thresholded recall and unthresholded overlap shows that the improvement appears in both better retrieval at a specified overlap threshold and higher average quality of the selected interval. 

The response differs for TR-DETR and Sim-DETR, with declines limited to 0.32--3.26 percentage points across the metrics. This variation may reflect a trade-off in the language input: Vocabulary-relative cooridinates supply sibling context through pooled description vectors, replacing the native sentence-token sequence. The reduction in explicit token-level detail may offset the benefit of this context for these two models.

These results give the vocabulary reference a second role in localization: 1) In D\'ej\`a Cue, it defines the query direction for scoring fixed candidates. 2) In learned models, it provides a language representation for the predictor that can improve temporal localization. The gains in models show that this representation can complement learned temporal modeling. The central connection is the use of the vocabulary to relate the queried state to its alternatives, whether the subsequent temporal selection is fixed or learned.

\section{Discussion}

State retrieval depends on the semantic reference used to read visual features as well as on the features themselves. Within an object history, high image--text similarity can reflect compatibility with the persistent object alongside evidence for its current state. The sibling vocabulary specifies the alternatives against which that evidence is interpreted. Since the state-balanced sibling average can vary across frames, changing the reference can reshape the temporal evidence used to compare intervals. The shared-candidate analysis links this semantic choice to temporal ranking: changing how observations are scored improves access to relevant intervals within the same search space. Localization gains across the tested encoders and text inputs support the broader relevance of this reference. Reference design thus complements feature extraction and temporal modeling by specifying which state distinctions the available features are asked to express.

The learned-model results suggest two complementary ways to carry this context into temporal prediction. Gains in four of the six architectures show that vocabulary-relative inputs can complement learned localization, while the differing responses motivate attention to how sibling context enters the predictor. One direction would retain native query-token features alongside centered sibling-description vectors, allowing a model to combine detail within each description with distinctions across states. A second would extend sibling conditioning from query construction to interval selection through joint ranking or decoding across the supplied states. This would allow relations between state descriptions to inform how candidate intervals are considered together. Both directions develop the same principle at two stages: representing language queries and selecting temporal intervals.

\vspace{-1em}

\paragraph{Limitations.} D\'ej\`a Cue uses tracked object histories and supplied state vocabularies. Discovering object identities and their state alternatives directly from video is a natural extension toward open-world state retrieval.

\section{Conclusion}
D\'ej\`a Cue uses a state-balanced vocabulary reference to read object states from frozen visual features. On MS-VOST, vocabulary-relative queries improve localization across encoder families and model sizes and promote useful intervals within a shared candidate set. Vocabulary-relative inputs also complement learning-based temporal localization models.

The vocabulary also frames the context of the retrieval problem. Uniform repetition within a state preserves the state-balanced reference, whereas altering the state set can shift query directions. Future extensions to discovered identities and evolving vocabularies would let both the object history and its semantic alternatives develop over time.

\clearpage
{
    \small
    \bibliographystyle{ieeenat_fullname}
    \bibliography{main}
}

\renewcommand{\maketitlesupplementary}{%
    \newpage
    \twocolumn[{%
        \centering
        {\Large\bfseries\thetitle\par}

        \vspace{1em}

        {\large
        \textbf{Code:}
        {\urlstyle{same}
        \href{https://github.com/HaofanCao/DejaCue}
        {https://github.com/HaofanCao/DejaCue}}
        \par}

        \vspace{1.6em}

        {\Large Supplementary Material\par}

        \vspace{1.4em}
    }]%
}

\clearpage
\setcounter{page}{1}
\maketitlesupplementary

\section{Detailed Method Derivations}
\label{sec:supp-method-derivations}

State-balanced text centering defines the vocabulary-relative query coordinate; trajectory centering supplies the dual variant. Both feed the same robust calibration and multiscale window statistic.

Indices $o$, $t$, $q$, and $s$ denote history, frame, description, and sibling state. The embedding dimension is $d$, the sequence length is $T$, $Q_o$ and $S_o$ count descriptions and sibling states, and $v_{o,t}$ indicates observation. Normalized visual and text embeddings are $\bar x_{o,t},\bar z_{o,q}\in\mathbb R^d$. Vectors are columns, $I_n$ is the $n\times n$ identity, and $\mathbf 1_n$ is the length-$n$ all-ones vector. Coordinate identities are algebraic; expectations and variances use only the stated working noise models.

\subsection{State-Balanced Origin and Projection Geometry}
\label{sec:8.1}
Let $\operatorname{st}_o(q)\in\{1,\ldots,S_o\}$ denote the sibling state containing description $q$. Assign each description the state-balanced weight
\begin{align}
\pi_{o,q}
&=\frac{1}{S_o|\mathcal Q_{o,\operatorname{st}_o(q)}|},\nonumber\\
\sum_{q=1}^{Q_o}\pi_{o,q}
&=\frac{1}{S_o}\sum_{s=1}^{S_o}
\frac{1}{|\mathcal Q_{o,s}|}
\sum_{q\in\mathcal Q_{o,s}}1=1.
\label{eq:supp-description-weight}
\end{align}
Because the state sets form a nonempty partition, the weights sum to one and each state carries total weight $1/S_o$.

The vocabulary origin is the weighted least-squares centroid. Define
\begin{align}
\mathcal J_o(c)
&=\sum_{q=1}^{Q_o}\pi_{o,q}\|\bar z_{o,q}-c\|_2^2,\nonumber\\
c_o&=\arg\min_{c\in\mathbb R^d}\mathcal J_o(c).
\label{eq:supp-centroid-objective}
\end{align}
Using $\sum_q\pi_{o,q}=1$, its gradient and Hessian are
\begin{align}
\nabla\mathcal J_o(c)
&=2\sum_{q=1}^{Q_o}\pi_{o,q}(c-\bar z_{o,q})
=2\!\left(c-\sum_{q=1}^{Q_o}\pi_{o,q}\bar z_{o,q}\right),\\
\nabla^2\mathcal J_o(c)&=2I_d.
\end{align}
The positive-definite Hessian makes $\mathcal J_o$ strictly convex, so its stationary point is the unique minimizer:
\begin{equation}
c_o=\sum_{q=1}^{Q_o}\pi_{o,q}\bar z_{o,q}
=\frac{1}{S_o}\sum_{s=1}^{S_o}
\frac{1}{|\mathcal Q_{o,s}|}\sum_{q\in\mathcal Q_{o,s}}\bar z_{o,q}.
\label{eq:supp-centroid-solution}
\end{equation}
The origin is therefore the arithmetic mean of the state prototypes, giving every sibling state equal influence.

A description-axis projection applies this centering to all descriptions at once. Collect the weights and normalized descriptions as
\begin{align}
\pi_o&=[\pi_{o,1},\ldots,\pi_{o,Q_o}]^{\top}
\in\mathbb R^{Q_o},\nonumber\\
\bar Z_o&=[\bar z_{o,1},\ldots,\bar z_{o,Q_o}]
\in\mathbb R^{d\times Q_o},\nonumber\\
H_o&=I_{Q_o}-\pi_o\mathbf 1_{Q_o}^{\top}.
\end{align}
Because $\bar Z_o\pi_o=c_o$, right multiplication by $H_o$ subtracts the same origin from every description:
\begin{equation}
\widetilde U_o
=\bar Z_oH_o
=\bar Z_o-(\bar Z_o\pi_o)\mathbf 1_{Q_o}^{\top}
=\bar Z_o-c_o\mathbf 1_{Q_o}^{\top}.
\label{eq:supp-residual-matrix}
\end{equation}
The $q$th column is the vocabulary-relative residual $\widetilde u_{o,q}=\bar z_{o,q}-c_o$. Using $\mathbf 1_{Q_o}^{\top}\pi_o=1$ gives
\begin{align}
H_o^2
&=I_{Q_o}-2\pi_o\mathbf 1_{Q_o}^{\top}
+\pi_o(\mathbf 1_{Q_o}^{\top}\pi_o)\mathbf 1_{Q_o}^{\top}
=H_o,\nonumber\\
H_o\pi_o&=0,
\qquad \mathbf 1_{Q_o}^{\top}H_o=0.
\label{eq:supp-centering-properties}
\end{align}
Thus $H_o$ is idempotent. It fixes every coefficient vector $a$ satisfying $\mathbf 1_{Q_o}^{\top}a=0$ and annihilates $\pi_o$, so its range is the zero-sum subspace and its kernel is $\operatorname{span}\{\pi_o\}$. It is generally an oblique, rather than orthogonal, projection because the state-balanced weights need not be uniform. The identity $\widetilde U_o\pi_o=0$ states that the weighted residual mean is zero.

Three exact consequences follow. First, for any $\delta\in\mathbb R^d$, adding the same offset to every normalized description has no effect:
\begin{equation}
(\bar Z_o+\delta\mathbf 1_{Q_o}^{\top})H_o
=\bar Z_oH_o+\delta(\mathbf 1_{Q_o}^{\top}H_o)
=\bar Z_oH_o.
\label{eq:supp-translation-invariance}
\end{equation}
Second, subtracting a common origin preserves all pairwise description differences before direction normalization:
\begin{equation}
(\bar z_{o,q}-c_o)-(\bar z_{o,j}-c_o)
=\bar z_{o,q}-\bar z_{o,j}.
\label{eq:supp-pairwise-preservation}
\end{equation}

Third, suppose every description in state $s$ is replicated uniformly $\rho_s\in\mathbb N_{+}$ times. The recomputed state mean is
\begin{equation}
\frac{1}{\rho_s|\mathcal Q_{o,s}|}
\sum_{q\in\mathcal Q_{o,s}}
\sum_{i=1}^{\rho_s}\bar z_{o,q}
=\mu^z_{o,s}.
\label{eq:supp-replication-invariance}
\end{equation}
Hence every state mean, $c_o$, and each original residual remain unchanged (\Cref{fig:duplication}). This invariance covers uniform replication within a state; genuinely new or asymmetrically added descriptions can change its mean.

At the state-prototype level, the corresponding centering operator is the orthogonal projector
\begin{equation}
\bar M_o=[\mu^z_{o,1},\ldots,\mu^z_{o,S_o}],
\qquad
P_o=I_{S_o}-S_o^{-1}\mathbf 1_{S_o}\mathbf 1_{S_o}^{\top}.
\end{equation}
Because $P_o^{\top}=P_o$ and $P_o^2=P_o$, it is the orthogonal projector onto the state-coefficient subspace $\{a\in\mathbb R^{S_o}:\mathbf 1_{S_o}^{\top}a=0\}$.
Its output $G_o=\bar M_oP_o$ is the unique closest prototype matrix whose columns sum to zero:
\begin{equation}
G_o=\arg\min_{G\in\mathbb R^{d\times S_o}}
\frac{1}{2}\|G-\bar M_o\|_F^2
\quad\text{s.t.}\quad G\mathbf 1_{S_o}=0.
\label{eq:supp-nearest-zero-sum}
\end{equation}
To verify the constrained optimum explicitly, introduce a multiplier $\lambda\in\mathbb R^d$ and the Lagrangian
\begin{equation}
\mathcal L(G,\lambda)
=\frac{1}{2}\|G-\bar M_o\|_F^2
+\lambda^{\top}G\mathbf 1_{S_o}.
\end{equation}
Stationarity and feasibility give
\begin{align}
G-\bar M_o+\lambda\mathbf 1_{S_o}^{\top}&=0,\nonumber\\
\lambda=S_o^{-1}\bar M_o\mathbf 1_{S_o}&=c_o,\nonumber\\
G=\bar M_o-c_o\mathbf 1_{S_o}^{\top}&=\bar M_oP_o.
\end{align}
Strict convexity on the affine feasible set makes this solution unique. The prototype-level orthogonal projection and description-level oblique projection share the same state-balanced origin $c_o$ but act on different column spaces.

Finally, let $y\in\mathbb R^d$ and define $a_{o,j}(y)=y^{\top}\bar z_{o,j}$. The unnormalized residual score is
\begin{align}
y^{\top}\widetilde u_{o,q}
&=a_{o,q}(y)-\sum_{j=1}^{Q_o}\pi_{o,j}a_{o,j}(y)\nonumber\\
&=a_{o,q}(y)-\frac{1}{S_o}\sum_{s=1}^{S_o}
\frac{1}{|\mathcal Q_{o,s}|}
\sum_{j\in\mathcal Q_{o,s}}a_{o,j}(y).
\label{eq:supp-contrast-score}
\end{align}
Thus the residual score compares query similarity with the state-balanced sibling average. Whenever $\|\widetilde u_{o,q}\|_2>\tau$, direction normalization divides this contrast by a positive number and preserves its sign. Residuals at or below $\tau$ instead use the absolute direction $\bar z_{o,q}$, avoiding division by a near-zero norm.

\subsection{Unified Coordinate Variants}

Let $\mathcal I_o=\{t:v_{o,t}=1\}$ be the nonempty set of observed frames. The visual origin $p_o$ is a coordinate-wise median, equivalently an $\ell_1$ location estimator:
\begin{equation}
p_o\in\arg\min_{p\in\mathbb R^d}
\sum_{t\in\mathcal I_o}\|\bar x_{o,t}-p\|_1,
\label{eq:supp-trajectory-l1}
\end{equation}
because, writing $\nu$ for a feature coordinate,
\begin{equation}
\sum_{t\in\mathcal I_o}\|\bar x_{o,t}-p\|_1
=\sum_{\nu=1}^{d}\sum_{t\in\mathcal I_o}
|\bar x_{o,t,\nu}-p_{\nu}|.
\end{equation}
Each scalar term is minimized when no more than half of the observations lie strictly on either side of $p_{\nu}$, which is precisely the median condition. When the sample count is even, any point in the scalar median interval is a valid minimizer; any fixed coordinate-wise median convention selects one such point.

The absolute and trajectory-centered visual coordinates are
\begin{equation}
r^{\mathrm{abs}}_{o,t}=\bar x_{o,t},\qquad
r^{\mathrm{traj}}_{o,t}=
\frac{\bar x_{o,t}-p_o}
{\max(\|\bar x_{o,t}-p_o\|_2,10^{-12})},
\end{equation}
respectively. The second vector has unit norm whenever $\|\bar x_{o,t}-p_o\|_2\geq10^{-12}$; below that numerical safeguard its norm is at most one, and it is exactly zero when $\bar x_{o,t}=p_o$.

For text, use the residual $\widetilde u_{o,q}=\bar z_{o,q}-c_o$ with threshold $\tau=10^{-8}$. The absolute and vocabulary-relative query coordinates are
\begin{align}
u^{\mathrm{abs}}_{o,q}&=\bar z_{o,q},\\
u^{\mathrm{vocab}}_{o,q}&=
\begin{cases}
\widetilde u_{o,q}/\|\widetilde u_{o,q}\|_2,
& \|\widetilde u_{o,q}\|_2>\tau,\\
\bar z_{o,q},&\text{otherwise}.
\end{cases}
\label{eq:supp-coordinate-definitions}
\end{align}
For $a\in\{\mathrm{abs},\mathrm{traj}\}$ and $b\in\{\mathrm{abs},\mathrm{vocab}\}$, all coordinate ablations use the same evidence function
\begin{equation}
e^{(a,b)}_{o,q,t}=(r^a_{o,t})^{\top}u^b_{o,q}.
\label{eq:supp-unified-evidence}
\end{equation}
The absolute baseline uses $(\mathrm{abs},\mathrm{abs})$. The trajectory-only ablation uses $(\mathrm{traj},\mathrm{abs})$, D\'ej\`a Cue uses the vocabulary-relative pair $(\mathrm{abs},\mathrm{vocab})$, and the dual variant uses $(\mathrm{traj},\mathrm{vocab})$. The D\'ej\`a Cue correspondence is
\begin{equation}
r_{o,t}=r^{\mathrm{abs}}_{o,t},\qquad
u_{o,q}=u^{\mathrm{vocab}}_{o,q},\qquad
e_{o,q,t}=e^{(\mathrm{abs},\mathrm{vocab})}_{o,q,t}.
\label{eq:supp-main-coordinate-correspondence}
\end{equation}
The trajectory-centered dual variant is
\begin{align}
r^{\mathrm{dual}}_{o,t}&=r^{\mathrm{traj}}_{o,t},\qquad
u^{\mathrm{dual}}_{o,q}=u^{\mathrm{vocab}}_{o,q},\nonumber\\
e^{\mathrm{dual}}_{o,q,t}
&=e^{(\mathrm{traj},\mathrm{vocab})}_{o,q,t}.
\label{eq:supp-dual-variant-correspondence}
\end{align}
Consequently, every coordinate comparison changes only the chosen visual and query origins; features, visibility, temporal candidates, calibration, and the downstream readout remain fixed. The downstream derivations use the unadorned core response $e_{o,q,t}$; they apply unchanged to the dual variant after substituting $e^{\mathrm{dual}}_{o,q,t}$.

\subsection{Robust Calibration and Local Support}
\label{sec:8.3}
For a fixed query $(o,q)$, calibration uses only its observed responses. To separate the MAD from its numerical floor, write
\begin{align}
m_{o,q}
&=\operatorname*{median}_{t\in\mathcal I_o}e_{o,q,t},\nonumber\\
\widehat s_{o,q}
&=1.4826\operatorname*{median}_{t\in\mathcal I_o}
|e_{o,q,t}-m_{o,q}|,\nonumber\\
s_{o,q}&=\max(\widehat s_{o,q},10^{-3}),\qquad
g_{o,q,t}=\frac{e_{o,q,t}-m_{o,q}}{s_{o,q}}.
\label{eq:supp-calibration-definitions}
\end{align}
The standardized response is invariant to any positive affine change when the scale floor is inactive. Let $e'_{o,q,t}=\alpha e_{o,q,t}+\beta$ with $\alpha>0$ and $\beta\in\mathbb R$, and let primed statistics be recomputed from $e'$. Positive affine equivariance of the median and absolute deviation gives
\begin{align}
m'_{o,q}&=\alpha m_{o,q}+\beta,\nonumber\\
|e'_{o,q,t}-m'_{o,q}|
&=\alpha|e_{o,q,t}-m_{o,q}|,\nonumber\\
\widehat s'_{o,q}&=\alpha\widehat s_{o,q}.
\end{align}
If $\widehat s_{o,q}>10^{-3}$ and $\alpha\widehat s_{o,q}>10^{-3}$, then $s_{o,q}=\widehat s_{o,q}$ and $s'_{o,q}=\alpha\widehat s_{o,q}$. Therefore
\begin{equation}
g'_{o,q,t}
=\frac{\alpha e_{o,q,t}+\beta-(\alpha m_{o,q}+\beta)}
{\alpha\widehat s_{o,q}}
=g_{o,q,t}.
\label{eq:supp-affine-invariance}
\end{equation}
The factor $1.4826$ calibrates the MAD to the Gaussian standard deviation. If $Z\sim\mathcal N(0,1)$ and $\Phi$ is the standard-normal cumulative distribution function, let $c_{\mathcal N}$ be the population median of $|Z|$. Then
\begin{align}
\Pr(|Z|\leq c_{\mathcal N})
=2\Phi(c_{\mathcal N})-1&=\frac{1}{2},\nonumber\\
c_{\mathcal N}&=\Phi^{-1}(0.75)\simeq0.67449.
\end{align}
Thus $1/c_{\mathcal N}\simeq1.4826$, making the scaled population MAD equal to the Gaussian standard deviation. The floor handles nearly constant responses and deliberately relaxes exact scale invariance only in that degenerate regime.

The local-support denominator preserves marginal noise variance even when the kernel contracts at a run boundary. For an odd support width $k$, a frame $t$ in observed run $R$, and radius $r_k=(k-1)/2$, define
\begin{align}
J_R^{(k)}(t)
&=\{j\in\mathbb Z:|j|\leq r_k,\ t+j\in R\},\nonumber\\
n_{R,t}&=|J_R^{(k)}(t)|\geq1.
\end{align}
Since the uniform weights are $h_j=1/k$ on the active offsets,
\begin{equation}
\begin{aligned}
\widetilde g_{o,q,t}
&=\frac{k^{-1}\sum_{j\in J_R^{(k)}(t)}g_{o,q,t+j}}
{\sqrt{n_{R,t}/k^2}}\\
&=\frac{1}{\sqrt{n_{R,t}}}
\sum_{j\in J_R^{(k)}(t)}g_{o,q,t+j}.
\end{aligned}
\label{eq:supp-uniform-support}
\end{equation}
Under the local working model $g_{o,q,t}=\mu+\xi_t$, the $\xi_t$ inside a run are independent with $\mathbb E[\xi_t]=0$ and $\operatorname{Var}(\xi_t)=1$. The aggregate moments are
\begin{align}
\mathbb E[\widetilde g_{o,q,t}]
&=\frac{1}{\sqrt{n_{R,t}}}
\sum_{j\in J_R^{(k)}(t)}\mu
=\mu\sqrt{n_{R,t}},\nonumber\\
\operatorname{Var}(\widetilde g_{o,q,t})
&=\frac{1}{n_{R,t}}
\sum_{j\in J_R^{(k)}(t)}\operatorname{Var}(\xi_{t+j})
=1.
\label{eq:supp-support-moments}
\end{align}
Thus a shorter boundary kernel accumulates less signal but does not receive an artificially smaller marginal noise variance merely because fewer taps are active. Because every offset in $J_R^{(k)}(t)$ remains inside the same maximal observed run, neither signal nor noise is propagated across a missing observation.

\subsection{Window Statistic, Duration Preference, and Complexity}

First isolate the duration normalization from smoothing-induced correlation. Fix a query $(o,q)$, let the target interval $w^\star$ of length $L$ lie inside one observed run, and consider a candidate $w$ in that run with length $\ell=|w|$. Under the independent-evidence working model
\begin{equation}
\widetilde g_{o,q,t}=\mu\mathbf 1[t\in w^\star]+\epsilon_t,
\qquad \mu>0,
\end{equation}
the residuals $\epsilon_t$ are independent, zero mean, and unit variance. This model is exact for $k=1$ and isolates duration normalization when $k>1$; the resulting short-range covariance is derived separately. Define the overlap length $\eta(w)=|w\cap w^\star|$. Substitution into the normalized window score gives
\begin{equation}
S(o,q,w)
=\frac{\mu\eta(w)}{\sqrt\ell}
+\frac{1}{\sqrt\ell}\sum_{t\in w}\epsilon_t.
\label{eq:supp-window-decomposition}
\end{equation}
Consequently,
\begin{align}
\mathbb E[S(o,q,w)]
&=\frac{\mu\eta(w)}{\sqrt\ell},\nonumber\\
\operatorname{Var}(S(o,q,w))
&=\frac{1}{\ell}\sum_{t\in w}\operatorname{Var}(\epsilon_t)=1.
\label{eq:supp-window-moments}
\end{align}
Since $\eta(w)\leq\min(\ell,L)$,
\begin{equation}
\mathbb E[S(o,q,w)]
\leq
\begin{cases}
\mu\sqrt\ell,&\ell\leq L,\\
\mu L/\sqrt\ell,&\ell\geq L.
\end{cases}
\label{eq:supp-duration-bound}
\end{equation}
The upper bound increases up to $\ell=L$ and decreases afterward, reaching $\mu\sqrt L$ only for a length-matched candidate with complete overlap. Square-root normalization therefore removes the linear preference for long sums while retaining the $\sqrt L$ gain from sustained evidence.

For $k>1$, adjacent locally aggregated responses share raw noise samples. Applying the working noise model to Equation~\ref{eq:supp-uniform-support} gives the noise component at any $t\in R$:
\begin{equation}
\zeta_t=\frac{1}{\sqrt{n_{R,t}}}
\sum_{j\in J_R^{(k)}(t)}\xi_{t+j}.
\end{equation}
For a fixed candidate $w=[a,b]$, let $\Sigma_w\in\mathbb R^{\ell\times\ell}$ be the covariance matrix of $(\zeta_a,\ldots,\zeta_b)$. The exact score variance is the variance of their normalized sum:
\begin{equation}
\operatorname{Var}(S(o,q,w))
=\frac{1}{\ell}\mathbf 1_{\ell}^{\top}
\Sigma_w\mathbf 1_{\ell}.
\label{eq:supp-correlated-window-variance}
\end{equation}
Away from run boundaries, $n_{R,t}=k$ and the active offsets are $-r_k,\ldots,r_k$. At temporal lag $\Delta\geq0$, the full-support sums for $\zeta_t$ and $\zeta_{t+\Delta}$ share $k-\Delta$ raw samples when $\Delta<k$ and none otherwise. Independence of the $\xi_t$ therefore yields
\begin{equation}
\gamma_{\Delta}
=\operatorname{Cov}(\zeta_t,\zeta_{t+\Delta})
=\begin{cases}
(k-\Delta)/k,&0\leq\Delta<k,\\
0,&\Delta\geq k.
\end{cases}
\label{eq:supp-local-autocovariance}
\end{equation}
In the full-support interior case, $\Sigma_w$ is Toeplitz with entries determined by Equation~\ref{eq:supp-local-autocovariance}. At each positive lag $\Delta$, its two off-diagonals contain $2(\ell-\Delta)$ entries, giving
\begin{equation}
\operatorname{Var}(S)
=1+\frac{2}{\ell}
\sum_{\Delta=1}^{\min(k-1,\ell-1)}
(\ell-\Delta)\frac{k-\Delta}{k}.
\label{eq:supp-uniform-window-variance}
\end{equation}
For fixed $k$, the large-window limit is
\begin{equation}
\lim_{\ell\to\infty}\operatorname{Var}(S)
=1+2\sum_{\Delta=1}^{k-1}\frac{k-\Delta}{k}
=k.
\end{equation}
Thus smoothing introduces short-range correlation, but the normalized score variance approaches a constant rather than growing with candidate duration. The covariance quadratic form remains exact near run boundaries, whereas the stationary Toeplitz simplification requires full support.

Finally, every score is evaluated in constant time after a cumulative sum. For run $R=[a_R,b_R]$, define the query-specific cumulative evidence
\begin{align}
C_{o,q,R}(t)&=\sum_{j=a_R}^{t}\widetilde g_{o,q,j},\\
S(o,q,[a,b])&=
\frac{C_{o,q,R}(b)-C_{o,q,R}(a-1)}
{\sqrt{b-a+1}},
\label{eq:supp-prefix-score}
\end{align}
for every candidate $[a,b]\subseteq R$, with $C_{o,q,R}(a_R-1)=0$. Starting from $\ell_1=1$, the duration grid follows
\begin{equation}
\ell_{m+1}=\min\!\left(T,\max\!\left\{\ell_m+1,
\operatorname{round}(1.5\ell_m)\right\}\right).
\label{eq:supp-duration-grid}
\end{equation}
Let $D$ be the number of generated lengths. Before the final cap at $T$, once $\ell_m\geq2$,
\begin{equation}
\ell_{m+1}
\geq\operatorname{round}(1.5\ell_m)
\geq1.5\ell_m-0.5
\geq\frac{5}{4}\ell_m.
\end{equation}
Hence $D\leq2+\lceil\log_{5/4}T\rceil=O(\log T)$. For a fixed length $\ell$, the number of valid starts across all observed runs is
\begin{equation}
N_{\ell}
=\sum_R\max\{|R|-\ell+1,0\}
\leq\sum_R|R|\leq T.
\end{equation}
Building all run-wise cumulative sums costs $O(T)$ time and memory, and enumerating every start at all $D$ lengths costs $O(TD)=O(T\log T)$ time per query. Because both cumulative sums and candidates are defined run by run, the missing-observation constraint is enforced by construction rather than by post-processing.

\section{Dataset and Evaluation}
\label{supp:data}

\paragraph{States and annotations.}

We construct MS-VOST by manually annotating object states in VOST videos. Two annotators independently label the frames. Each state is described by a short label and a visual definition, which serve as separate queries for the same state. An object history contains its observed frame sequence and the vocabulary of its three to five states. The dataset comprises 620 histories from 620 source videos, 2,465 states, 4,930 descriptions, and 2,473 annotated occurrences (Table~\ref{tab:dataset}).

\begin{table}[htbp]
\centering
\caption{\textbf{MS-VOST composition.} Occurrences are annotated temporal intervals. Each state provides a label and a visual definition as two distinct queries.}
\label{tab:dataset}
\small
\setlength{\tabcolsep}{3pt}
\begin{tabular}{@{}lr@{}}
\toprule
Quantity & Count \\
\midrule
Source videos / object histories & 620 / 620 \\
States & 2,465 \\
Descriptions & 4,930 \\
Annotated occurrences & 2,473 \\
Native video frames & 582,705 \\
Histories with 3 / 4 / 5 states & 18 / 599 / 3 \\
States with multiple occurrences & 6 \\
\bottomrule
\end{tabular}
\end{table}

\begin{table*}[htbp]
\centering
\caption{\textbf{Frozen encoder configurations.} Pixels denotes input resolution and Dim denotes embedding dimension.}
\vspace{-0.5em}
\label{tab:checkpoints}
\small
\setlength{\tabcolsep}{3pt}
\begin{tabular}{@{}llrr@{}}
\toprule
Encoder & Checkpoint & Pixels & Dim \\
\midrule
CLIP B/16 & \texttt{openai/clip-vit-base-patch16} & 224 & 512 \\
SigLIP1 B/16 & \texttt{google/siglip-base-patch16-224} & 224 & 768 \\
MetaCLIP B/16 & \texttt{facebook/metaclip-b16-fullcc2.5b} & 224 & 512 \\
SigLIP2 B/16 & \texttt{google/siglip2-base-patch16-224} & 224 & 768 \\
SigLIP2 B/16 & \texttt{google/siglip2-base-patch16-256} & 256 & 768 \\
SigLIP2 L/16 & \texttt{google/siglip2-large-patch16-256} & 256 & 1024 \\
SigLIP2 So400m/16 & \texttt{google/siglip2-so400m-patch16-256} & 256 & 1152 \\
SigLIP2 Giant-opt/16 & \texttt{google/siglip2-giant-opt-patch16-256} & 256 & 1536 \\
\bottomrule
\end{tabular}
\vspace{-0.5em}
\end{table*}

For each occurrence, the annotations specify a stable interval and plausible outer boundaries. The stable interval runs from the first definite start to the last definite end. The plausible boundaries mark the earliest plausible start and latest plausible end. These annotations define the four reference choices evaluated in the main paper: Stable, Earlier start, Later end, and Broader. Repeated occurrences retain separate intervals associated with the same state. Annotation times are mapped to the nearest native video timestamps; ties select the earlier frame. All intervals use inclusive frame endpoints.

\paragraph{Object observations.}

All methods use the same target masks and object crops. Grounding DINO initializes the target from its description, and SAM~2.1 Hiera-Large propagates the masks. We correct missed or incorrect tracks and combine parts belonging to a split target. We expand the target bounding box on each side by 20\% of its larger dimension, with a minimum of four pixels and clipping at the image boundary. Pixels outside the target mask are set to RGB~127. Each encoder applies its corresponding image processor to this crop. A nonempty mask marks a visible observation; invisible frames retain their original indices.

\vspace{-0.5em}

\paragraph{Metrics and averaging.}

For a prediction $w_{o,q}$ and the annotated intervals $\mathcal{R}_{o,s(q)}$ of its state, define $\mathrm{IoU}_{o,q}=\max_{r\in\mathcal{R}_{o,s(q)}}\mathrm{tIoU}(w_{o,q},r)$. Temporal IoU uses inclusive interval lengths. The per-query values are $\mathrm{IoU}_{o,q}$ for Top-1 tIoU and $\mathbf{1}[\mathrm{IoU}_{o,q}\geq\theta]$ for R@1 at threshold $\theta\in\{0.3,0.5\}$. Descriptions are averaged within states, states within histories, and histories within source videos; source videos receive equal weight. Score differences are computed before rounding.

Candidate-ranking measures use the same averaging. Candidates are sorted by score, then by earlier start and earlier end to resolve ties. At a tIoU threshold of 0.5, reciprocal rank is the inverse rank of the first relevant interval, or zero if none exists. R@$K$ records whether at least one of the first $K$ candidates is relevant. The candidate oracle is the largest available tIoU for the query, independent of candidate order. These definitions distinguish the intervals offered by search from their ordering under each coordinate system.

\section{Training-free Implementation}
\label{supp:implementation}

Table~\ref{tab:checkpoints} identifies the frozen encoders. All use $16\times16$ patches. The family comparison holds image resolution at 224 pixels; the SigLIP2 size comparison uses 256 pixels. The reference encoder for component and learned-model studies is SigLIP2 Base/16 at 224 pixels. Image features and each text-form feature are unit normalized. For the default text representation, we encode the description $d$, ``\textit{a photo of }$d$", and ``\textit{the }$d$", average their three normalized embeddings, and normalize the average.

\paragraph{Coordinate construction.}

For each history, we first average description embeddings within each state and then average the state prototypes. This state-balanced mean is the vocabulary origin. D\'ej\`a Cue subtracts that origin from each description embedding and normalizes the residual; its visual embeddings remain uncentered. If the residual norm is at most $10^{-8}$, the normalized original description is used. Absolute uses the uncentered image and text embeddings. Dual additionally subtracts the coordinate-wise median of the visible visual embeddings and normalizes the visual residuals.

\vspace{-0.5em}

\paragraph{Temporal retrieval.}

Each query produces one response per visible frame. We subtract the median response and divide by the larger of $1.4826$ times the median absolute deviation and $10^{-3}$. For an even number of observations, a median averages the two central values. Three-frame smoothing is applied separately within each contiguous visible interval. Only available in-interval taps enter the weighted sum and its normalization, as derived in \Cref{sec:8.3}.

Candidate lengths follow the integer geometric schedule $\ell_1=1$ and $\ell_{j+1}=\min\{T,\max[\ell_j+1,\operatorname{round}(1.5\ell_j)]\}$, stopping at the native history length $T$. Rounding uses the nearest integer, with ties to the even integer. Every fitting length is evaluated at every start within a contiguous visible interval. A window receives the sum of its smoothed responses divided by the square root of its length. The maximum-scoring window is returned, with ties resolved by earlier start and then earlier end. Coordinate comparisons share this entire temporal procedure.

\section{State Weighting and the Dual Extension}
\label{supp:construction}

\subsection{Does description repetition change reference?}
\label{sec:11.1}
Uniformly repeating the descriptions of one state should not increase that state\textquotesingle s contribution to the vocabulary origin. We examine this property by repeating both descriptions of the first state in each history by factors of two, four, and eight. Predictions for the original 4,930 descriptions are compared with their predictions before repetition. The experiment uses Dual for all eight encoders and changes only the weighting used to construct the vocabulary mean.

\begin{figure*}[htbp]
    \centering
    \includegraphics[width=0.87\linewidth]{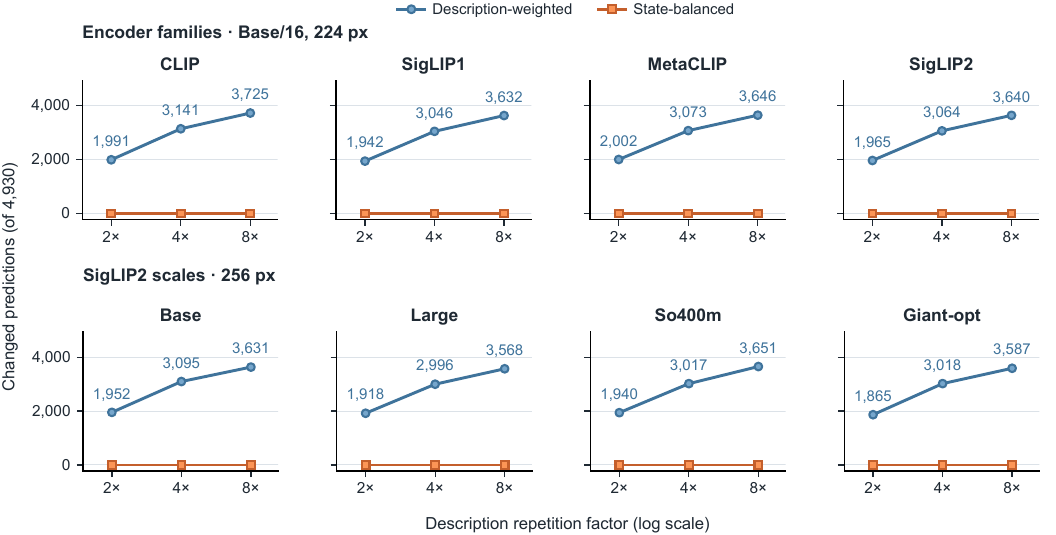}
    \vspace{-0.3em}
    \caption{\textbf{Prediction stability under repeated descriptions.} Both descriptions of the first state in each history are repeated by the indicated factor. Curves count changed predictions among the original 4,930 queries under Dual coordinates; all panels share the same scale.}
    \vspace{-0.5em}
    \label{fig:duplication}
\end{figure*}

\begin{figure*}[htbp]
    \centering
\includegraphics[width=0.87\linewidth]{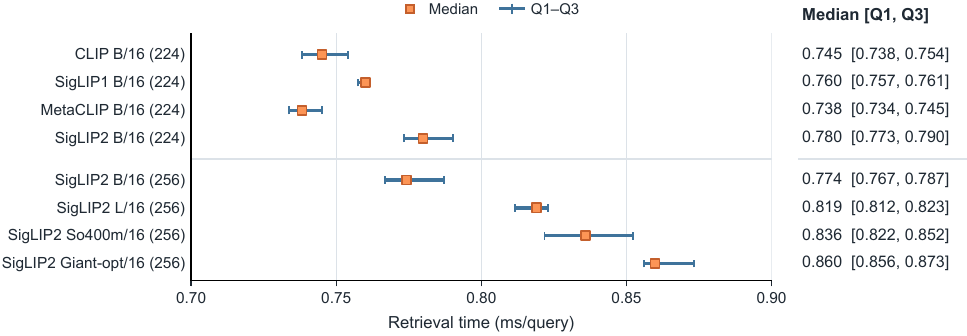}
\vspace{-0.4em}
    \caption{\textbf{Dual retrieval time across encoders.} Squares show medians and horizontal ranges show the 25th--75th percentiles (Q1--Q3) across ten timed passes on an RTX~4090D, using precomputed features. Values at right give median [Q1, Q3] in milliseconds per query.}
    \vspace{-1em}
\label{fig:timing}
\end{figure*}

\Cref{fig:duplication} shows the number of changed intervals. State-balanced weighting preserves every original prediction for all eight encoders and all three repetition factors. Equal weighting of descriptions changes 1,965, 3,064, and 3,640 predictions for the reference encoder. This is the retrieval consequence of the identity in \Cref{sec:8.1}: uniform replication leaves the state prototype, vocabulary origin, and query direction unchanged. The reference therefore reflects the represented states independently of how many times their descriptions are listed.

\subsection{What does the visual reference add?}

The vocabulary origin and the visual trajectory provide complementary reference choices. Table~\ref{tab:dual} reports Absolute, D\'ej\`a Cue, and Dual under the same temporal search for all eight encoders. Dual combines the state-balanced vocabulary origin with the visual median described in \Cref{supp:implementation}. Relative to Absolute, its Top-1 tIoU gains range from 0.63 to 1.49 percentage points. Relative to vocabulary-only D\'ej\`a Cue, the additional Top-1 tIoU gain is 0.20--0.37 points. This extension retains semantic reference and also expresses each observation relative to its own trajectory.

\vspace{-1.2em}

\begin{table*}[htbp]
\centering
\caption{\textbf{Vocabulary-relative coordinates only and Dual coordinates on MS-VOST.} Scores are percentages.}
\vspace{-0.5em}
\label{tab:dual}
\small
\setlength{\tabcolsep}{2.5pt}
\resizebox{\linewidth}{!}{%
\begin{tabular}{@{}cllrr*{12}{r}@{}}
\toprule
& & & & &
\multicolumn{4}{c}{R@1@0.3 $\uparrow$} &
\multicolumn{4}{c}{R@1@0.5 $\uparrow$} &
\multicolumn{4}{c}{Top-1 tIoU $\uparrow$} \\
\cmidrule(lr){6-9}
\cmidrule(lr){10-13}
\cmidrule(lr){14-17}
& Encoder & Scale & Pixels & Dim
& Abs. & D\'ej\`a & Dual & $\Delta$
& Abs. & D\'ej\`a & Dual & $\Delta$
& Abs. & D\'ej\`a & Dual & $\Delta$ \\
\midrule
\multirow{4}{*}{\rotatebox[origin=c]{90}{\shortstack{Encoder\\families}}}
& CLIP \citep{radford2021clip} & Base & 224 & 512
& $13.34$ & $14.40$ & \textbf{14.91} & +1.56
& $6.31$ & $6.94$ & \textbf{7.11} & +0.80
& $9.90$ & $10.59$ & \textbf{10.91} & +1.01 \\

& SigLIP1 \citep{zhai2023siglip} & Base & 224 & 768
& $13.76$ & \textbf{14.44} & $14.32$ & +0.56
& $6.42$ & $6.82$ & \textbf{6.87} & +0.45
& $10.17$ & $10.59$ & \textbf{10.80} & +0.63 \\

& MetaCLIP \citep{xu2024metaclip} & Base & 224 & 512
& $13.47$ & $14.73$ & \textbf{15.30} & +1.83
& $6.26$ & $6.89$ & \textbf{7.09} & +0.83
& $10.12$ & $10.95$ & \textbf{11.19} & +1.07 \\

& SigLIP2 \citep{tschannen2025siglip2} & Base & 224 & 768
& $13.27$ & $14.54$ & \textbf{14.73} & +1.47
& $6.44$ & $7.14$ & \textbf{7.18} & +0.73
& $10.01$ & $10.78$ & \textbf{10.99} & +0.98 \\

\midrule
\multirow{4}{*}{\rotatebox[origin=c]{90}{\shortstack{SigLIP2\\scales}}}
& SigLIP2 \citep{tschannen2025siglip2} & Base & 256 & 768
& $13.28$ & $14.38$ & \textbf{15.11} & +1.83
& $6.58$ & $7.06$ & \textbf{7.25} & +0.68
& $10.09$ & $10.82$ & \textbf{11.11} & +1.02 \\

& SigLIP2 \citep{tschannen2025siglip2} & Large & 256 & 1024
& $13.08$ & $14.82$ & \textbf{15.19} & +2.11
& $6.17$ & $7.00$ & \textbf{7.04} & +0.86
& $9.98$ & $10.82$ & \textbf{11.04} & +1.06 \\

& SigLIP2 \citep{tschannen2025siglip2} & So400m & 256 & 1152
& $13.88$ & $14.88$ & \textbf{15.43} & +1.55
& $6.30$ & $7.27$ & \textbf{7.33} & +1.03
& $10.14$ & $11.27$ & \textbf{11.47} & +1.33 \\

& SigLIP2 \citep{tschannen2025siglip2} & Giant-opt & 256 & 1536
& $13.24$ & $15.26$ & \textbf{15.77} & +2.53
& $6.42$ & $7.44$ & \textbf{7.49} & +1.07
& $9.93$ & $11.04$ & \textbf{11.41} & +1.49 \\

\bottomrule
\end{tabular}%
}
\par\vspace{3pt}
\begin{minipage}{\linewidth}
\footnotesize
\textit{Note.} Abs. (Absolute) uses uncentered embeddings; D\'ej\`a (D\'ej\`a Cue) centers the vocabulary; Dual also centers the visual trajectory.
Scores are percentages; $\Delta$ denoted Dual minus Abs. (pp).
Pixels denotes input resolution; Dim denotes embedding dimension. Differences are computed before rounding.
\end{minipage}
\vspace{-0.5em}
\end{table*}

\paragraph{Retrieval cost.}

To characterize the cost of this extension, we time Dual retrieval with precomputed image and text features on an NVIDIA RTX~4090D. Each pass processes all 620 histories and 4,930 queries. After 3 warm-up passes, ten timed passes give median retrieval times of 0.738--0.860 ms per query across the eight encoders (\Cref{fig:timing}). The measurement covers coordinate construction, response processing, and temporal search; feature extraction and mask generation precede timed stage. All 8 configurations remain below one millisecond per query for this retrieval stage.

\begin{figure*}[htbp]
    \centering
    \includegraphics[width=0.85\linewidth]{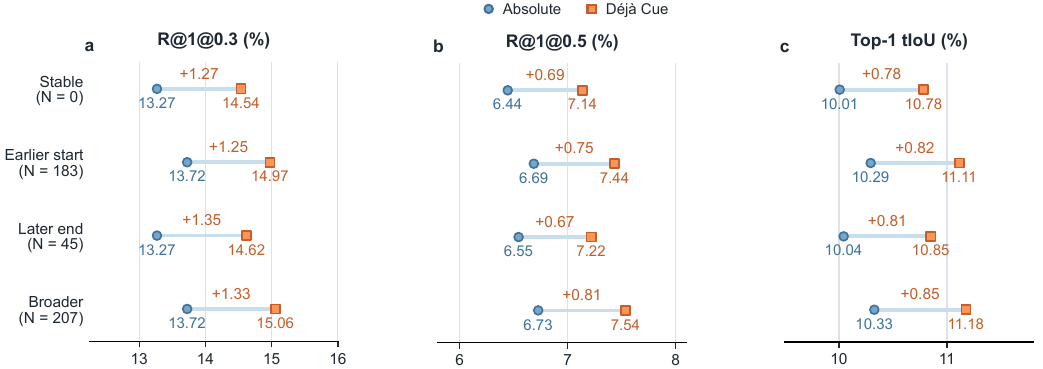}
    \vspace{-0.5em}
    \caption{\textbf{Localization under annotated boundary choices on MS-VOST.} Points show Absolute and D\'ej\`a Cue scores with fixed predictions. N counts intervals whose endpoint change; all 2,473 intervals (i.e. 4,930 queries) are scored under every definition.}
    \vspace{-0.5em}
    \label{fig:boundaries}
\end{figure*}

\section{Additional Localization Results}
\label{supp:analyses}

\subsection{Text Forms across Encoders}

Tables~\ref{tab:text-inputs} (in the end of this supplementary) give the three localization metrics underlying the main text-input analysis. Each row compares Absolute and  D\'ej\`a Cue with the same encoder, visual features, text form, and temporal search. Raw, Photo, and Article denote $d$, ``\textit{a photo of }$d$", and ``\textit{the }$d$", respectively. Average is the normalized mean of the three unit-normalized embeddings.

All three metrics increase in each of the 32 encoder--input configurations. Top-1 tIoU gains range from 0.20 to 1.34 percentage points, including positive gains for every individual text form. The improvement therefore persists when the reference is built from a single wording convention as well as from the default average. Together with the encoder comparison, these results place the vocabulary reference across the tested representation families, model sizes, and text-feature constructions.

\begin{table*}[!t]
\centering
\caption{\textbf{Localization across state durations on MS-VOST.} Scores are percentages.}
\vspace{-0.5em}
\label{tab:duration}
\small
\setlength{\tabcolsep}{3pt}
\begin{tabular}{@{}lr*{9}{r}@{}}
\toprule
& &
\multicolumn{3}{c}{R@1@0.3 $\uparrow$} &
\multicolumn{3}{c}{R@1@0.5 $\uparrow$} &
\multicolumn{3}{c}{Top-1 tIoU $\uparrow$} \\
\cmidrule(lr){3-5}
\cmidrule(lr){6-8}
\cmidrule(lr){9-11}
Duration & $n$
& Abs. & D\'ej\`a & $\Delta$
& Abs. & D\'ej\`a & $\Delta$
& Abs. & D\'ej\`a & $\Delta$ \\
\midrule

$\leq 1.13$ s & 619
& $3.80$ & $4.12$ & \textbf{+0.32}
& $1.20$ & $1.92$ & \textbf{+0.72}
& $2.68$ & $3.36$ & \textbf{+0.68} \\

$1.13$--$2.87$ s & 615
& $10.58$ & $13.83$ & \textbf{+3.25}
& $5.78$ & $6.99$ & \textbf{+1.20}
& $7.76$ & $9.66$ & \textbf{+1.90} \\

$2.87$--$7.03$ s & 615
& $18.45$ & $20.31$ & \textbf{+1.86}
& $10.20$ & $10.95$ & \textbf{+0.76}
& $13.33$ & $14.11$ & \textbf{+0.78} \\

$>7.03$ s & 616
& $19.36$ & $18.67$ & --0.70
& $7.46$ & $6.93$ & --0.53
& $15.79$ & $15.45$ & --0.33 \\

\bottomrule
\end{tabular}

\par\vspace{3pt}
\begin{minipage}{0.62\linewidth}
\footnotesize\textit{Note.} Duration is the mean annotated occurrence length of a state; $n$ counts states. Both methods use SigLIP2 Base/16 at 224 pixels and the same temporal search. $\Delta$ is D\'ej\`a Cue minus Absolute.
\end{minipage}
\vspace{-1.2em}
\end{table*}

\subsection{Annotated Boundaries and State Duration}

The boundary analysis scores the same predictions against four annotated reference definitions. Earlier start extends only the stable onset; Later end extends only the stable offset; Broader extends both. \Cref{fig:boundaries} gives all three metrics. The number of intervals with changed endpoints is 183 for Earlier start, 45 for Later end, and 207 for Broader, including 21 intervals with both boundaries extended.

D\'ej\`a Cue improves all three metrics under each reference definition, with Top-1 tIoU gains of 0.78--0.85 percentage points. The changes in absolute scores reflect the adjusted annotation boundaries, while the positive coordinate gains persist across these interpretations of state onset and offset. This supports the boundary analysis in the main paper without changing the retrieved intervals.

Table~\ref{tab:duration} supplies the thresholded-recall results alongside the Top-1 tIoU values shown in the main duration figure. States are grouped by their mean annotated occurrence duration, with quartile boundaries of 1.1333, 2.8667, and 7.0333 seconds. The four groups contain 619, 615, 615, and 616 states. Each group uses the same nested averaging.

The first three groups improve across all three metrics, with the largest Top-1 tIoU gain in the 1.13--2.87-second group. For the longest states, Top-1 tIoU decreases by 0.33 percentage points, while both coordinate systems attain their highest overlap in this group. Thus, absolute overlap and the gain from reference vary differently with duration: longer states have higher overlap, whereas the largest coordinate benefit appears at an intermediate duration.

\subsection{Reading a Four-state History}

\begin{figure}[!h]
\centering
\includegraphics[width=\linewidth]{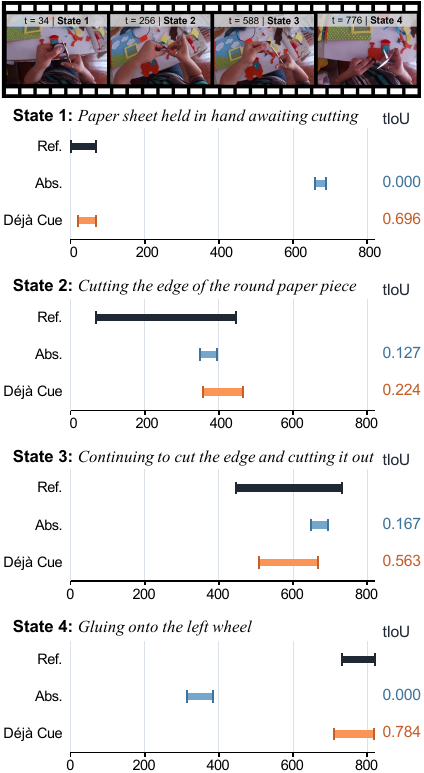}\caption{\textbf{Reading four states within one MS-VOST history.} Panels use each state\textquotesingle s label as the query. Sampled frames accompany the Reference interval, Absolute, and D\'ej\`a Cue predictions from SigLIP2 Base/16 at 224 pixels. All panels share the frame axis; tIoU is shown at right.}
\label{fig:history}
\end{figure}

Figure~\ref{fig:history} follows all four annotated states of a paper-cutting and assembly history. Each state\textquotesingle s short label is used as the query, with SigLIP2 Base/16 at 224 pixels. The four panels share the native frame axis and show the saved predictions from the same temporal search. For the initial paper state, D\'ej\`a Cue moves the prediction from frames 658--689 to 21--68, within the reference interval of 0--68. For gluing the piece onto the wheel, it moves from 314--385 to 711--818, closely overlapping the reference interval of 732--821.

The two intermediate states distinguish successive stages of cutting the same paper piece. D\'ej\`a Cue increases their temporal overlap from 0.127 to 0.224 and from 0.167 to 0.563, respectively. For the latter query, the retrieved interval shifts earlier and lengthens from 647--694 to 507--668, covering more of the annotated cutting stage at 445--732. Across this history, the reference supports both corrections to the selected temporal region and broader coverage within it. The sampled frames provide visual context, while the aligned intervals show which part of the observation sequence each query retrieves.
\vspace{-0.3em}

\section{Inputs for Learned Models}
\label{supp:learned}
\vspace{-0.2em}

The learned-model comparison asks whether a temporal predictor can use the supplied sibling descriptions as its language context. Figure~\ref{fig:learned-input} shows the two input constructions. Original receives the token features of the queried sentence. Vocabulary-Relative Coordinates (VRC) representation receives one pooled vector for every description in the history, including the queried description, after state-balanced centering and residual normalization. A binary channel identifies the queried description. With SigLIP2 Base/16, these inputs have 768 channels (i.e. $d$ = 768) for Original and 769 for VRC before the model\textquotesingle s input projection.

\clearpage

\begin{strip}
\centering
\includegraphics[width=\linewidth]{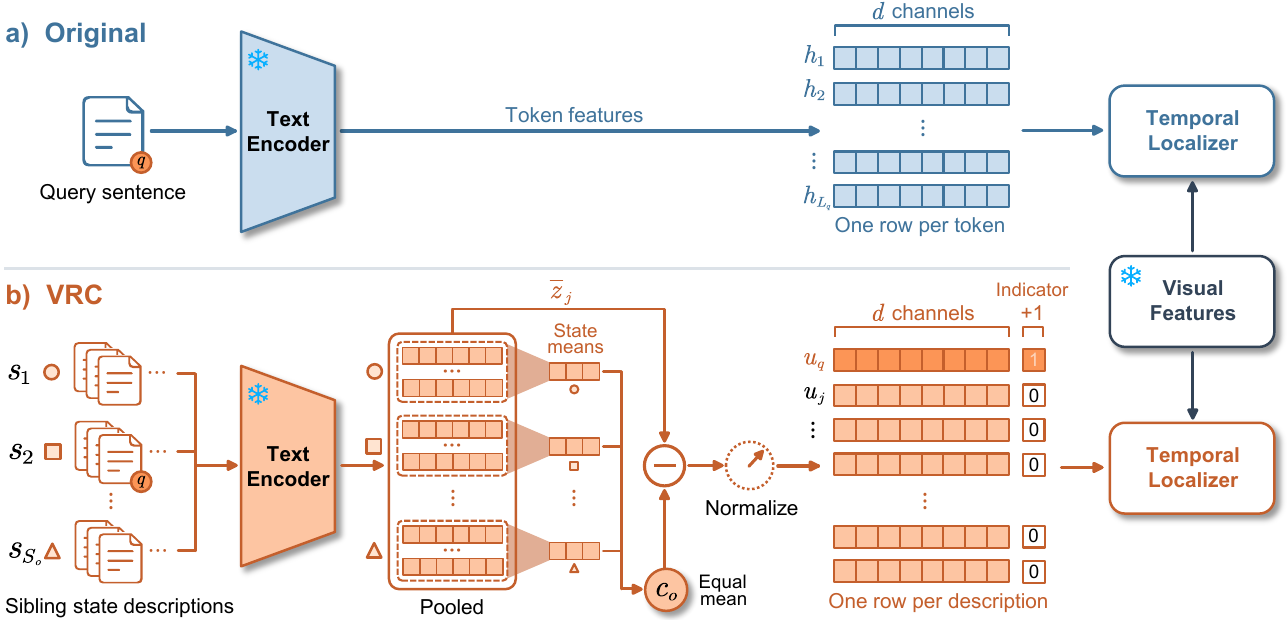}
\captionsetup{hypcap=false}
\captionof{figure}{%
\textbf{Language inputs for learned localization.}
(a) Original represents the query as $L_q$ token features
$h_i \in \mathbb{R}^d$.
(b) VRC centers all pooled sibling-description vectors on the
state-balanced mean $c_o$, normalizes the residuals, and appends
a query-indicator channel to each vector.
Both conditions use the same frozen visual features.
Model-specific sequence ordering is described in the text.%
}
\label{fig:learned-input}
\captionsetup{hypcap=true}
\end{strip}

For query $q$, the VRC representation contains $[u_{o,j};\mathbf{1}(j=q)]$ for every description $j$. The queried description is placed first and the other descriptions retain their original order. Sim-DETR places the queried description at the last valid position, moving its indicator with it. Padding is masked. This construction presents the queried state together with the descriptions that define its alternatives; the representation replaces the native sentence-token sequence.

\paragraph{Training and scoring.}

The five folds are grouped by source video and shared across methods and input conditions (Table~\ref{tab:folds}). Each run trains on four folds and predicts the remaining fold. We use seeds 3407, 3408, and 3409, with 200 epochs per run. Both input conditions use the frozen SigLIP2 Base/16 encoder, AdamW with initial learning rate and weight decay of $10^{-4}$, and gradient-norm clipping at 0.1. For each architecture, Original and VRC are trained separately. For each seed, the five evaluation-fold predictions are combined before scoring. The reported mean and sample standard deviation are computed from the resulting three seed scores.

\begin{table}[htbp]
\centering
\caption{\textbf{Source-video folds for learned localization.} Each fold is used once for evaluation and the other four for training. Counts describe the evaluation fold.}
\label{tab:folds}
\small
\setlength{\tabcolsep}{3pt}
\begin{tabular}{@{}lrrr@{}}
\toprule
Fold & Histories & States & Queries \\
\midrule
1 & 125 & 498 & 996 \\
2 & 124 & 494 & 988 \\
3 & 124 & 490 & 980 \\
4 & 123 & 486 & 972 \\
5 & 124 & 497 & 994 \\
\bottomrule
\end{tabular}
\vspace{-0.5em}
\end{table}

\paragraph{Visible intervals.}

A training example is a contiguous visible interval with nonempty overlap with a reference occurrence; target spans are the intersections of those occurrences with that interval. At evaluation, the model processes every visible interval, and proposals are ranked jointly for the query. Predicted spans are mapped to the original frame indices. TaskWeave uniformly samples intervals longer than 75 frames to 75 positions and pads shorter intervals, retaining the original time coordinates when mapping its output back to the history.

To sum up, these details clarify the common role of the vocabulary across the two forms of localization. In the training-free method, the vocabulary determines the query direction used to score fixed candidates. In the learned setting, it determines a language representation that the predictor can use alongside the visual sequence. The main part of this paper reports the six-model comparison and its architecture-dependent response (\Cref{sec:5.4}); the present construction here specifies the contextual input underlying that comparison.

\clearpage

\begin{table*}[!b]
\centering
\vspace{1em}
\caption{\textbf{Text-input comparison across encoder families and SigLIP2 scales on MS-VOST.} Scores are percentages.}
\label{tab:text-inputs}
\small
\setlength{\tabcolsep}{2.5pt}
\begin{tabular}{@{}ccl*{3}{rcr}@{}}
\toprule
& & & \multicolumn{3}{c}{R@1@0.3 $\uparrow$} & \multicolumn{3}{c}{R@1@0.5 $\uparrow$} & \multicolumn{3}{c}{Top-1 tIoU $\uparrow$} \\
\cmidrule(lr){4-6}\cmidrule(lr){7-9}\cmidrule(lr){10-12}
& Encoder & Text & Absolute & D\'ej\`a Cue & $\Delta$ & Absolute & D\'ej\`a Cue & $\Delta$ & Absolute & D\'ej\`a Cue & $\Delta$ \\
\midrule

\multirow{16}{*}{\rotatebox[origin=c]{90}{\shortstack{\normalsize Encoder families}}}
& \multirow{4}{*}{\rotatebox[origin=c]{90}{\shortstack{CLIP\\Base}}}
& Raw & $13.61$ & $14.01$ & \textbf{+0.40} & $6.41$ & $6.88$ & \textbf{+0.47} & $9.89$ & $10.57$ & \textbf{+0.67} \\
& & Photo & $13.61$ & $14.87$ & \textbf{+1.26} & $6.40$ & $7.11$ & \textbf{+0.71} & $9.95$ & $10.82$ & \textbf{+0.87} \\
& & Article & $13.36$ & $14.44$ & \textbf{+1.09} & $6.32$ & $6.87$ & \textbf{+0.55} & $9.84$ & $10.59$ & \textbf{+0.75} \\
& & Avg. & $13.34$ & $14.40$ & \textbf{+1.06} & $6.31$ & $6.94$ & \textbf{+0.64} & $9.90$ & $10.59$ & \textbf{+0.68} \\

\cmidrule(lr){2-12}

& \multirow{4}{*}{\rotatebox[origin=c]{90}{\shortstack{SigLIP1\\Base}}}
& Raw & $13.76$ & $14.55$ & \textbf{+0.79} & $6.32$ & $6.93$ & \textbf{+0.61} & $10.15$ & $10.68$ & \textbf{+0.53} \\
& & Photo & $14.15$ & $14.63$ & \textbf{+0.48} & $6.58$ & $6.81$ & \textbf{+0.23} & $10.44$ & $10.64$ & \textbf{+0.20} \\
& & Article & $13.53$ & $14.56$ & \textbf{+1.02} & $6.41$ & $6.85$ & \textbf{+0.44} & $10.16$ & $10.55$ & \textbf{+0.39} \\
& & Avg. & $13.76$ & $14.44$ & \textbf{+0.68} & $6.42$ & $6.82$ & \textbf{+0.40} & $10.17$ & $10.59$ & \textbf{+0.42} \\

\cmidrule(lr){2-12}

& \multirow{4}{*}{\rotatebox[origin=c]{90}{\shortstack{MetaCLIP\\Base}}}
& Raw & $13.89$ & $15.27$ & \textbf{+1.38} & $6.45$ & $7.19$ & \textbf{+0.74} & $10.28$ & $11.05$ & \textbf{+0.77} \\
& & Photo & $13.22$ & $14.83$ & \textbf{+1.61} & $6.15$ & $6.87$ & \textbf{+0.72} & $9.99$ & $10.95$ & \textbf{+0.96} \\
& & Article & $13.71$ & $14.89$ & \textbf{+1.19} & $6.38$ & $6.97$ & \textbf{+0.58} & $10.12$ & $10.77$ & \textbf{+0.65} \\
& & Avg. & $13.47$ & $14.73$ & \textbf{+1.26} & $6.26$ & $6.89$ & \textbf{+0.63} & $10.12$ & $10.95$ & \textbf{+0.82} \\

\cmidrule(lr){2-12}

& \multirow{4}{*}{\rotatebox[origin=c]{90}{\shortstack{SigLIP2\\Base}}}
& Raw & $13.53$ & $14.49$ & \textbf{+0.96} & $6.36$ & $6.90$ & \textbf{+0.53} & $10.18$ & $10.67$ & \textbf{+0.49} \\
& & Photo & $13.68$ & $15.31$ & \textbf{+1.63} & $6.73$ & $7.15$ & \textbf{+0.42} & $10.26$ & $10.98$ & \textbf{+0.72} \\
& & Article & $13.51$ & $14.65$ & \textbf{+1.15} & $6.41$ & $7.26$ & \textbf{+0.85} & $10.11$ & $10.83$ & \textbf{+0.72} \\
& & Avg. & $13.27$ & $14.54$ & \textbf{+1.27} & $6.44$ & $7.14$ & \textbf{+0.69} & $10.01$ & $10.78$ & \textbf{+0.78} \\

\midrule

\multirow{16}{*}{\rotatebox[origin=c]{90}{\shortstack{\normalsize SigLIP2 scales}}}
& \multirow{4}{*}{\rotatebox[origin=c]{90}{\shortstack{SigLIP2\\Base}}}
& Raw & $13.33$ & $14.56$ & \textbf{+1.24} & $6.28$ & $6.46$ & \textbf{+0.18} & $10.02$ & $10.61$ & \textbf{+0.59} \\
& & Photo & $13.46$ & $15.02$ & \textbf{+1.56} & $6.64$ & $7.29$ & \textbf{+0.65} & $10.11$ & $11.03$ & \textbf{+0.92} \\
& & Article & $13.15$ & $14.47$ & \textbf{+1.32} & $6.12$ & $7.35$ & \textbf{+1.23} & $9.93$ & $10.90$ & \textbf{+0.97} \\
& & Avg. & $13.28$ & $14.38$ & \textbf{+1.10} & $6.58$ & $7.06$ & \textbf{+0.49} & $10.09$ & $10.82$ & \textbf{+0.73} \\

\cmidrule(lr){2-12}

& \multirow{4}{*}{\rotatebox[origin=c]{90}{\shortstack{SigLIP2\\Large}}}
& Raw & $12.85$ & $14.57$ & \textbf{+1.72} & $6.09$ & $6.71$ & \textbf{+0.61} & $9.86$ & $10.69$ & \textbf{+0.84} \\
& & Photo & $13.32$ & $14.76$ & \textbf{+1.44} & $6.33$ & $6.95$ & \textbf{+0.62} & $10.07$ & $10.77$ & \textbf{+0.70} \\
& & Article & $12.55$ & $14.71$ & \textbf{+2.16} & $6.20$ & $7.01$ & \textbf{+0.81} & $9.72$ & $10.72$ & \textbf{+1.01} \\
& & Avg. & $13.08$ & $14.82$ & \textbf{+1.74} & $6.17$ & $7.00$ & \textbf{+0.82} & $9.98$ & $10.82$ & \textbf{+0.84} \\

\cmidrule(lr){2-12}

& \multirow{4}{*}{\rotatebox[origin=c]{90}{\shortstack{SigLIP2\\So400m}}}
& Raw & $13.63$ & $15.13$ & \textbf{+1.50} & $6.32$ & $7.38$ & \textbf{+1.06} & $10.09$ & $11.27$ & \textbf{+1.18} \\
& & Photo & $13.77$ & $15.26$ & \textbf{+1.49} & $6.01$ & $7.57$ & \textbf{+1.56} & $10.01$ & $11.35$ & \textbf{+1.34} \\
& & Article & $13.23$ & $14.77$ & \textbf{+1.54} & $6.17$ & $7.29$ & \textbf{+1.12} & $9.89$ & $11.05$ & \textbf{+1.16} \\
& & Avg. & $13.88$ & $14.88$ & \textbf{+1.00} & $6.30$ & $7.27$ & \textbf{+0.97} & $10.14$ & $11.27$ & \textbf{+1.14} \\

\cmidrule(lr){2-12}

& \multirow{4}{*}{\rotatebox[origin=c]{90}{\shortstack{SigLIP2\\Giant-opt}}}
& Raw & $13.20$ & $14.71$ & \textbf{+1.51} & $6.49$ & $7.27$ & \textbf{+0.78} & $9.97$ & $10.89$ & \textbf{+0.92} \\
& & Photo & $13.40$ & $15.12$ & \textbf{+1.72} & $6.55$ & $7.19$ & \textbf{+0.64} & $10.00$ & $11.01$ & \textbf{+1.00} \\
& & Article & $12.79$ & $14.82$ & \textbf{+2.03} & $6.17$ & $6.73$ & \textbf{+0.56} & $9.70$ & $10.76$ & \textbf{+1.06} \\
& & Avg. & $13.24$ & $15.26$ & \textbf{+2.02} & $6.42$ & $7.44$ & \textbf{+1.01} & $9.93$ & $11.04$ & \textbf{+1.12} \\

\bottomrule
\end{tabular}

\par\vspace{3pt}
\begin{minipage}{0.77\linewidth}
\footnotesize
\textit{Note.} Absolute uses uncentered embeddings; D\'ej\`a Cue centers the vocabulary. Raw, Photo, and Article encode $d$, ``\textit{a photo of }$d$", and ``\textit{the }$d$"; Avg. normalizes their averaged embeddings. The upper group uses Base/16 encoders at 224 pixels; the lower group uses SigLIP2 sizes with 16$\times$16 patches at 256 pixels. Differences are computed before rounding.
\end{minipage}
\end{table*}

\section{On the Naming of D\'ej\`a Cue}
\label{sec:supp-naming}

\vspace{-0.5em}

The term \emph{D\'ej\`a vu} (French for ``already seen'') describes the psychological illusion of feeling broadly familiar with a situation that is actually being experienced for the first time. In the context of persistent object histories, however, this familiarity is not an illusion: the tracked object is constantly present across the video. As discussed in our formulation, this persistent identity can make many observed moments look broadly familiar to any query mentioning the object, causing absolute image--text similarity to conflate state-specific evidence with general object compatibility.

We therefore call our method \textbf{D\'ej\`a Cue}. While the tracked history provides the \emph{D\'ej\`a}---the already seen, persistent identity that establishes the foundational context---it is not enough to simply recognize the object. The state-balanced sibling vocabulary provides the critical \emph{Cue}: the relative direction of support needed to distinguish exactly which semantic state the visual evidence depicts. Rather than being overwhelmed by the baseline familiarity of the object, our method isolates the definitive cue among its alternatives.

\end{document}